\documentclass[acmlarge, nonacm]{acmart}

\AtBeginDocument{%
  \providecommand\BibTeX{{%
    \normalfont B\kern-0.5em{\scshape i\kern-0.25em b}\kern-0.8em\TeX}}}

\usepackage{amsmath}
\usepackage{balance}       
\usepackage{graphics}      
\usepackage[T1]{fontenc}   
\usepackage{lipsum}
\usepackage{color}
\usepackage{booktabs}
\usepackage{arydshln}
\usepackage{tabularray}
\usepackage{booktabs}
\usepackage{arydshln}
\usepackage{textcomp}
\usepackage{microtype}        
\usepackage{ccicons}          
\usepackage{multirow}  
\usepackage{color,soul}  
\usepackage{subcaption}  
\usepackage{todonotes}
\usepackage{colortbl}  
\usepackage{adjustbox}
\usepackage[normalem]{ulem}
\usepackage{float}
\usepackage{graphicx}
\usepackage{multirow}
\usepackage{tabularx}
\usepackage{longtable}

\usepackage{nicematrix,tikz}

\NiceMatrixOptions
  {
    custom-line = 
     {
       letter = : ,
       command = dashedline , 
       ccommand = cdashedline ,
       tikz = dashed
     }
  }

\definecolor{ao(english)}{rgb}{0.0, 0.5, 0.0}
\definecolor{awesome}{rgb}{1.0, 0.13, 0.32}

\usepackage{array}
\newcolumntype{P}[1]{>{\centering\arraybackslash}p{#1}}

\usepackage{fancyhdr}
\begin{document}
 \thispagestyle{fancy}
\fancyhead{} 
\fancyfoot{} 
\pagenumbering{gobble}
\fancyfoot[C]{\textcolor{red}{This manuscript is under review. Please write to mthukral3@gatech.edu or sourish.dhekane@gatech.edu  for up-to-date information}}
\title[Layout Agnostic HAR in Smart Homes through TDOST]{Layout Agnostic Human Activity Recognition in Smart Homes through \underline{T}extual \underline{D}escriptions \underline{O}f \underline{S}ensor \underline{T}riggers (TDOST)}

\author{Megha Thukral}
\authornote{Both authors contributed equally to this research.}
\email{mthukral3@gatech.edu}
\orcid{1234-5678-9012}
\author{Sourish Gunesh Dhekane}
\authornotemark[1]
\email{sourish.dhekane@gatech.edu}
\affiliation{%
	\institution{School of Interactive Computing, Georgia Institute of Technology}
	\city{Atlanta, GA}
	\country{USA}
}
\author{Shruthi K.  Hiremath}
\email{shiremath9@gatech.edu}
\affiliation{
	\institution{School of Interactive Computing, Georgia Institute of Technology}
	\city{Atlanta, GA}
	\country{USA}
}
\author{Harish Haresamudram}
\email{hharesamudram3@gatech.edu}
\affiliation{%
	\institution{School of Electrical and Computer Engineering, Georgia Institute of Technology}
	\city{Atlanta, GA}
	\country{USA}
}

\author{Thomas Pl\"{o}tz}
\email{thomas.ploetz@gatech.edu}
\affiliation{
	\institution{School of Interactive Computing, Georgia Institute of Technology}
	\city{Atlanta, GA}
	\country{USA}
}

\renewcommand{\shortauthors}{Thukral, et al.}

\begin{abstract}
\input{abstract.tex}
\end{abstract}


\begin{abstract}
  Human activity recognition (HAR) using ambient sensors in smart homes has numerous applications for human healthcare and wellness. 
  However, building general-purpose HAR models that can be deployed to new smart home environments requires a significant amount of annotated sensor data and training overhead. 
  Most smart homes vary significantly in their layouts, i.e., floor plans and the specifics of sensors embedded, resulting in low generalizability of HAR models trained for specific homes.
  We address this limitation by introducing a novel, layout-agnostic modeling approach for HAR systems in smart homes that utilizes the transferrable representational capacity of natural language descriptions of raw sensor data.
  To this end, we generate \textit{Textual Descriptions Of Sensor Triggers (TDOST)} that encapsulate the surrounding trigger conditions and provide cues for underlying activities to the activity recognition models. 
  Leveraging textual embeddings, rather than raw sensor data, we create activity recognition systems that predict standard activities across homes without either (re-)training or adaptation on target homes.
  Through an extensive evaluation, we demonstrate the effectiveness of TDOST-based models in unseen smart homes through experiments on benchmarked CASAS datasets. 
  Furthermore, we conduct a detailed analysis of how the individual components of our approach affect downstream activity recognition performance.
\end{abstract}
\maketitle

\section{Introduction}
\label{sec:introduction}

Human Activity Recognition (HAR) using ambient sensors in smart homes is crucial for numerous healthcare and monitoring applications \cite{kientz2008georgia, helal2005gator, chan2008review, alam2012review}. 
The ubiquitous nature of these sensors provides a non-intrusive, low-cost, and privacy-preserving modality that can be deployed in people's living spaces and effectively be used for activity recognition.
State-of-the-art approaches  \cite{ghods2019activity2vec,bouchabou2021fully, bouchabou2021using, liciotti2020sequential, bouchabou2021human, chen2023leveraging, singh2017convolutional} have achieved usable activity recognition accuracy by deploying deep neural networks that, however, rely on large amounts of labeled training data for the predominant supervised learning approach.
Even worse, these requirements apply for each and every single individual home as their layouts (floorplans) and sensor instrumentation typically differ substantially from the original training scenario.

This raises the practical issue that models trained for one--source--home cannot easily be deployed ``as is'' in other--target--homes.  
One way to overcome this problem is to retrain the HAR model with labeled sensor data collected for each target home.
For real-world applications, however, the accompanying high computational and financial resource demands for such a straightforward procedure seem infeasible.

In this work, we propose a novel approach for deriving \emph{layout-agnostic HAR} systems for common sets of activities that perform  well in target homes without requiring (re-)training or adaptation on labeled sample data from a specific target home.
To be easily deployable in real-world homes, these HAR systems must perform well while being \textit{indifferent} to the underlying floor plan of the target house and the sensor arrangement from which the sensor activity is generated. 
We hypothesize that generalizable HAR systems that require no re-training with target data are achievable through the incorporation of contextual information that accompany the raw sensor triggers and can typically be obtained with minimal effort. 
Such contextual information includes but is not limited to the position of the sensor in the environment, its modality, and the time of the sensor trigger.

To utilize the aforementioned additional context, we construct natural language descriptions of sensor triggers that are enriched with context in which they were triggered: \textit{\underline{T}extual \underline{D}escriptions \underline{O}f \underline{S}ensor \underline{T}riggers (TDOST)}. 
TDOSTs can effectively abstract from the intricacies of the layout of a house and its sensor triggers using natural language sentences. 
They can also be efficiently encoded using pre-trained language models \cite{zhou2020evaluating, rajani2019explain}.
We demonstrate that the textual embedding space of web-scale trained text encoders can be used in HAR systems to transfer models across home layouts \emph{without the need for re-training on target layouts.}

Our approach addresses a range of challenges \cite{dhekane2024transfer} that arise when  state-of-the-art HAR models are deployed in new conditions. 
Specifically, we tackle the following using HAR models trained on TDOSTs: 
(\textit{i}) changes in sensor locations;
(\textit{ii})  different residents;
(\textit{iii}) need for data efficiency for model training; and
(\textit{iv}) need for label efficiency specifically w.r.t.\   target homes where no labeled or unlabeled data are collected.
To this end, we develop a HAR system that does not collect any training data, resident information, or activity labels from a target home.
Instead, our method only requires the floor plan of the target house and various meta-information such as quantity and types of sensors present in the house.

This information is \textit{only collected once} from the target home and is ingested automatically by our processing pipeline to convert raw sensor triggers to TDOSTs. 
The TDOSTs are then directly fed into a deep classifier that was trained only on labeled data from a source home. 
We emphasise that we do not retrain this model in target smart homes, thereby \textit{eliminating the need to collect dense activity annotations} in new smart homes.

Through a series of experimental evaluations, we demonstrate the capabilities of our approach on a range of benchmark smart home datasets. 
First, the TDOST HAR models are evaluated in the same source setting on which they are trained. 
Our layout-agnostic HAR model performs better or at par with the state-of the-art supervised baselines. 
This establishes the recognition capabilities of the TDOST-trained activity classifier.
More importantly, we then show the exceptional transfer capabilities of TDOST.
Without requiring retraining and not even adaptation, the recognition capabilities of our TDOST based models trained on individual source homes and transferred to target homes 
achieve superior recognition performance when compared to using state-of-the art procedures for the same transfer across given homes.

The main contributions of this paper can be summarized as follows: 
\begin{itemize}
    \item We present TDOST -- an approach that facilitates layout-agnostic HAR through encoding readily available contextual information into textual descriptions of sensor readings. Coarse-level layout information and sensor characteristics are the only information required for TDOST-based layout agnostic HAR.
    TDOST leverages large-scale pre-trained language models to build general-purpose models that can be transferred across target homes.

    \item Our approach is both data- and label-efficient showing state-of-the art performance on new smart homes without access to any target data be it labeled or unlabeled. 

    \item Through extensive experimentation, we study the robustness of our approach by deploying it in various scenarios based on standard benchmark datasets from CASAS \cite{cook2012casas}.
    We also evaluate the effectiveness of different variants of our proposed textual descriptions of sensor triggers.

\end{itemize}

\section{Related Work}
\label{sec:related}
Our work presents an approach for layout-agnostic human activity recognition (HAR) in smart homes.
We do so through the main component of TDOST, our proposed approach -- that uses natural language to describe sensor event triggers recorded in a given home. 
Using knowledge learnt through observing data in a source home, we deploy a HAR system in a target home, without requiring access to any data from the target home. 

In what follows, we summarize HAR systems as they are currently developed for smart homes, and discuss the various components, relevant to our work.

\subsection{Human Activity Recognition in Smart Homes}
\label{sec2:related-work:HAR}
With decreasing costs of ambient sensors and an increase in demand for home automation, instrumenting smart homes has become a reality for many \cite{cook2003mavhome, hooper2012french, alemdar2013aras, kientz2008georgia, helal2005gator}.
Yet, analyzing the data collected through the domotic sensors remains a challenging endeavor. 
Typically, activity recognition systems for smart homes are built with the goal of identifying activities and behaviors that residents engage in. 
Such systems are useful to log behaviors, identify routines and are, thus, applied in daily monitoring and  health related scenarios \cite{dang2020sensor, kulsoom2022review, chatting2023automated, morita2023health}. 

Aforementioned activity recognition systems can be built using either \textit{i)} knowledge-driven procedures or \textit{ii)} data-driven procedures as identified in \cite{bouchabou2021survey}. 
The former typically requires information from individual smart homes, the behaviors of resident(s), and interactions between the resident and the environment \cite{bouchabou2021survey, kulsoom2022review}. 
Ontologies need to be constructed that either use taxonomies of activities developed for specific homes \cite{chen2009ontology} or are bootstrapped to identify the context of specific activities \cite{boovaraghavan2023tao}. 
Thus, these require domain expertise, context knowledge and data collected in individual smart homes over extended periods of time.
In data-driven procedures the activity recognition systems comprise of two steps -- \textit{i)} segmentation and  \textit{ii)} classification, aimed at recognizing the activity label for an identified segment where the activity label corresponds to an activity class of interest. 
Segmentation approaches have been designed to use likelihood ratios, density metrics or statistical ratios to identify the contiguous segments of sensor data \cite{aminikhanghahi2017using, aminikhanghahi2018real}. 
Time-based or sensor-event based windowing procedures are also widely adopted, where in windows, typically assumed to be independently and identically distributed are identified \cite{quigley2018comparative}. 
In order to identify the appropriate length of windows, data needs to be collected in specific homes and analyzed. 
Given the variations across different homes, a one-size-fits-all windowing approach is typically not feasible \cite{allik2019optimization, fida2014effect, fida2015varying}. 

A number of traditional approaches has been employed in order to recognize the activity of interest from the identified segments. 
These use manually generated features \cite{cook2020activity}, which are then classified using approaches such as KNN, HMM, Naive Bayes, etc.  \cite{sedky2018evaluating, cook2010learning, fahad2014activity}. 
Utilizing spatio-temporal information \cite{bouchabou2021survey, chen2019human},  or context based information such as location or time duration \cite{hoque2012aalo, fahad2014activity} is a popular technique in the analysis of data collected in smart homes. 
The approach of cluster-then-classify usually makes use of such contextual information to aggregate sensor events that belong to a specific activity of interest \cite{hoque2012aalo, wu2022cluster, fahad2014activity}. 

More recently deep-learning based approaches have been employed for HAR in smart homes. 
In \cite{liciotti2020sequential}, authors use somewhat simplistic feature engineering yet showcase high recognition performance through employing (variants of) LSTMs. 
In \cite{bouchabou2021fully} and \cite{bouchabou2021using} language based embedding techniques have been used to learn mappings from sensor event triggers to an embedding space that identifies similarities between activity segments that belong to the same activity class. 
Graph neural network based approaches \cite{zhou2020graph,li2019relation,li2019relation} and self-supervision based techniques \cite{chen2024enhancing} have been employed to learn a better embedding space. 

Aiming to identify a good embedding space, we use textual descriptions for sensor event triggers. 
These descriptions map the knowledge obtained from sensor triggers in a source home to new target homes.

\subsection{Resources for HAR systems in Smart Homes:}
\label{sec2:related-work:Resources-HAR}
Developing robust activity recognition systems for specific homes is resource intensive in terms of data collected and annotations for the data collected \cite{intille2005placelab, hooper2012french, alemdar2013aras, cook2003mavhome}. 
Annotations are required to learn a mapping between the learned representations and activities of interest. 
Active-learning based procedures \cite{settles2009active} where annotations are self-provided by residents prompted at appropriate and opportune times \cite{rashidi2009keeping,intille2004acquiring} still rely on measures such as information gain or entropy to identify data points to query \cite{adaimi2019leveraging, dahmen2017activity}. 
Utilizing semi-supervision based techniques work in \cite{cook2013activity} used automated class discovery to identify informative data points.

Obtaining these resources is an expensive endeavor as shown in \cite{hiremath2020deriving, zhai2019s4l}. 
Work in \cite{hoque2012aalo} clusters data points collected in the home based on location and queries for identified clusters. 
In \cite{al2020zero} a zero-shot learning procedure is used to identify the closest activity label based on generated word embeddings.
These embeddings are generated using expert knowledge
and require domain expertise in identifying relevant context information or correspondences between similar activities. 
For instance, in \cite{al2020zero}, authors use a mapping between activities, e.g., `read' and `watch\_tv', in one home to the activity of `relax' in another home, thus requiring domain knowledge corresponding to both homes.  

In our approach, we aim to reduce the dependency on extensive data collection, obtaining annotations or requiring domain expertise to develop HAR systems.  

\subsection{Human Activity Recognition Systems for Generalizable Deployments: Technical Foundations}
\label{sec:related:new-home-HAR}

Developing activity recognition systems for individual smart homes is potentially cumbersome and expensive, requiring labeled data instances to train classification models. 
Thus, transfer learning or domain adaptation procedures have been employed to utilize knowledge gained in a specific source house (`domain') and to apply that to a target house (`domain'). 
More formally, the source domain $D_s$ and target $D_t$ domains differ from each other in one or more of the following: 
\textit{i)} underlying feature sets, $\chi_s \neq \chi_t$; and/or
\textit{ii)} marginal distributions of data points, $P(X_s) \neq P(X_t)$ or \textit{iii)} different label sets $Y_s \neq Y_t$ in source and target domains \cite{weiss2016survey, pan2009survey, torrey2010transfer, zhuang2020comprehensive}.
When label sets are the same across domains, $Y_s = Y_t$, the desired generalization of deployment of a HAR system becomes a domain adaptation problem \cite{farahani2021brief, csurka2017domain, chang2020systematic}.

Transfer learning is a paradigm where knowledge learned from one task (source task) is used to boost the performance on a related task (target task), by bridging the gap between the two \cite{hosna2022transfer}. 
Transfer learning methods are aimed at: 
\textit{i}) instance transfer; 
\textit{ii}) feature space transfer; and 
\textit{iii})  parameter transfer \cite{zhuang2020comprehensive, agarwal2021transfer}. 
Instance transfer methods transfer knowledge from the source domain in terms of data instances \cite{zhao2020makes}, where the goal is to address the data distribution shift across domains. 
Some of the approaches employed for instance transfer include instance weighting \cite{wang2019instance} -- where samples in the target domain are weighted based on the confidence of the predictions obtained from the source model, source domain selection \cite{bascol2019improving} where appropriate source domains are chosen to transfer knowledge and instance label mapping \cite{day2017survey}  usually employed to address the distribution shifts between the source and target domain instances.

More recently, instance transfer has also been performed by artificially generating the data instances in the target domain using the source data samples \cite{rotem2022transfer, li2024synthetic}.

Feature space transfer methods try to construct a feature space common to both the source and the target domains \cite{pan2009survey}.
Sensor profiling \cite{hu2011transfer, chiang2012knowledge}, sparse coding \cite{maurer2013sparse,chang2017unsupervised}, using genetic algorithms \cite{koccer2010genetic}, 
ontological methods \cite{sanabria2021unsupervised}, employing meta-features \cite{sun2019meta}, feature similarity metrics \cite{qin2019cross} are some of the methods used to develop feature transfer methods. 
Feature space transfer can also be performed using neural networks, which can accommodate heterogeneous data and project it on a common feature space.  
The heterogeneous sensor data, is passed through an embedding layer that projects it into the common feature space. 
Network architectures like autoencoders (AE), graph AE, long short term memory cells (LSTMs), and generative adversarial networks (GANs) have been used to obtain these embeddings. 
The network based feature space transfer methods can also be perceived as parameter transfer methods, where the parameters learned over the source domain (in terms of network weights) are reused in target domain inference \cite{he2021towards}. 

Similar to aforementioned motivations for adaptation,  changes across smart homes in sensor modalities, sensor arrangements, and different residents induce a difference in feature sets, $\chi_s \neq \chi_t$. 
Activity sets across the different homes may also be different depending on the idiosyncratic behaviors of residents of these homes. 
Our work on developing new recognition systems for homes differs from the traditional transfer learning settings \cite{khan2018untran, raina2007self, guo2018deep}.
Specifically, we assume access to no data instances from the target home,  to deploy an effective HAR system.

\label{sec:method}

\begin{figure}[t]
  \centering
  \includegraphics[width=\linewidth]{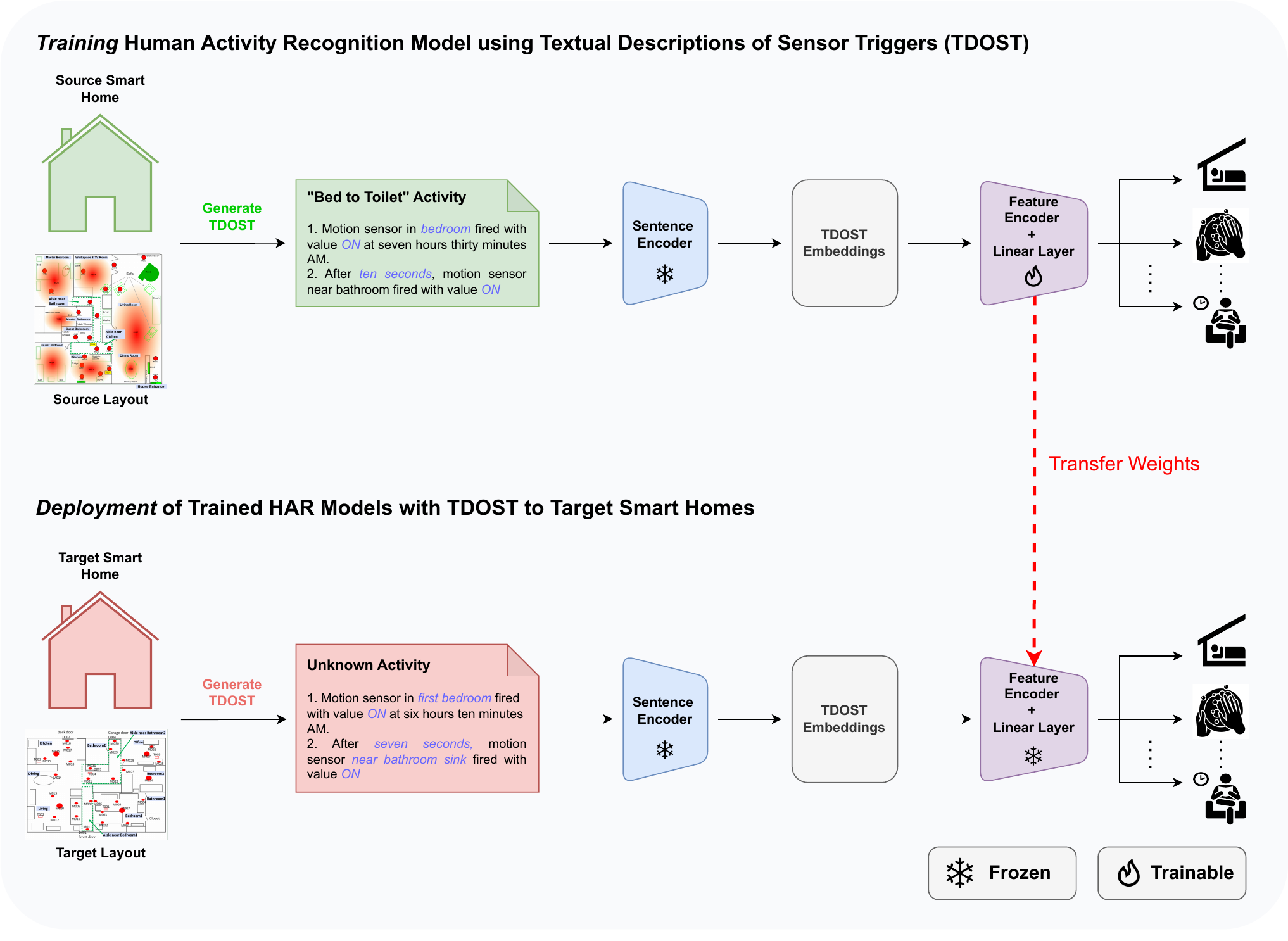} 
  \caption{Overview of layout-agnostic HAR: Our approach derives HAR models on some given source home that can be deployed in new target homes while being agnostic at the modeling level to changes in overall floor plans. 
  At the heart of our approach lies the construction of textual descriptions of raw sensor triggers--TDOST--encoding readily available context information such as symbolic sensor locations and sensor modalities. 
  Layout-agnostic HAR works in two phases: 
  (i) Supervised training of a HAR model in a source home using labeled activity data (top part),
  and (ii) Deployment of the trained source model to new target homes (bottom part). 
  Source data is first converted to textual descriptions which are ingested by frozen pre-trained sentence encoders. 
  The source embeddings are further adapted to smart home activity using a feature encoder and classification linear layer.
  To predict the standard set of activities (common across smart homes) in a new home, the trained source model is deployed ``as is'' without any adaptation/re-training. 
  During the inference stage in target homes, the test activity sensor data is converted to textual description in the same way as originally done in the source smart home -- thereby utilizing easy to obtain meta-information such as floorplan and sensor specifications. 
  The embeddings for target TDOST, generated from sentence encoders are further transformed using the feature encoder and classifier head trained in source. 
  During inference all models including sentence encoder, feature encoder, and classification head are kept frozen.
  }

\label{fig:method}
\end{figure}
\section{Layout Agnostic Human Activity Recognition in Smart Homes through Textual Description of Sensor Triggers --TDOST}

Deriving HAR systems for each individual smart home comes with high data collection efforts, extensive training, and annotation overhead. 
In order to alleviate such often impractical demands, we propose a novel approach that allows for transferring smart home HAR systems for standard sets of activities across homes thereby effectively abstracting from floorplan details and as such minimizing the manual effort required for HAR system transfer. 

At the heart of our approach are  \underline{T}extual \underline{D}escriptions \underline{O}f \underline{S}ensor \underline{T}riggers (TDOST) through which we convert raw sensor data into textual sentences.
We map these TDOST to a textual embedding space using a pre-trained sentence encoder. 
These sentence representations are then adapted for smart home activities by deploying a feature encoder and a linear classification layer. 
After training the feature encoder and classification layer with data from a source home, they can be utilized for activity recognition on standard activities in any new smart home without the need for re-training or annotation in the target homes. 
From target homes, we require the symbolic sensor locations, their types, and floor plans which are utilized in our inference pipeline to construct contextualized textual descriptions for target sensor triggers.
Figure \ref{fig:method} provides an overview of our approach, which we will describe in detail in this section.

\subsection{Generating Textual Descriptions for Smart Home Datasets:}\label{TDOST-Variants}

We hypothesize that a HAR model that can transfer across multiple smart homes must be trained with a significant amount of contextualized information in addition to raw sensor data.
State-of-the-art models focus on mapping sensor data to discrete symbols which are then used to train feature encoders. \cite{liciotti2020sequential}\cite{bouchabou2021using} 
However, this line of work does not leverage the contextual information associated with smart home sensors.
For instance, a motion sensor with a value ``ON'' would be mapped to an arbitrary integer without incorporating any information detailing its location or modality. 
An alternate representation that captures the necessary contextual cues would abstract from unnecessary technical details like sensor identity or overly precise absolute timestamps and thus would be more useful for downstream activity recognition.

To this end, we propose constructing natural text descriptions of sensor data (TDOST) that include contextual information solely derived from the smart home layout, i.e., floorplan and sensor placement, along with specifications of high-level sensor characteristics such as type of sensor.
Using textual descriptions has two benefits:
First, additional context information can be added in a straightforward manner to each sensor event using natural language and generic background information that can be obtained from external as well as internal sources, e.g., the internet and a house's floorplan, respectively.  
Yet, more importantly, the constructed sentences can be effectively encoded using pre-trained language models as they have been shown to exhibit common-sense capabilities \cite{zhou2020evaluating, rajani2019explain}.
In addition, these text models also provide a representation space that can be used to transfer across home layouts without the need for any re-training.

To derive informative textual descriptions, we consider diverse sources of context such as sensor location, type, and temporal information.
Sensor location and type provide crucial context about the potential activities occurring in the home. For example, a motion sensor triggered in the kitchen can be potentially linked to ``meal preparation'' activity.
Similarly, the time lag between two sensor triggers captures potential cues for the underlying activities. 
Overall, we introduce four variants of TDOST that capture context at varying granularity, namely:
\begin{enumerate}
    \item \textit{TDOST-Basic:} Captures location context and type of the sensor trigger.
    
    \item \textit{TDOST-Temporal:} Captures time lag between two sensor triggers in addition to sensor location and type.
    
    \item \textit{TDOST-LLM:} Contains multiple textual descriptions generated by GPT-4\cite{achiam2023gpt} for each sensor trigger.
    \item \textit{TDOST-LLM+Temporal:} Adds relative temporal information between sensor triggers to the sentences generated by GPT-4.
\end{enumerate}

\subsubsection{TDOST-Basic:} 
In this variant, through the following template, we add context to the raw sensor triggers in the form of sensor type / modality and symbolic sensor location in the house (which room), along with the encoded sensor value:

\begin{quote}
\texttt{<Sensor-Type> sensor in <Location of Sensor> fired with value <Sensor-Value>}
\end{quote}


Using this template, we convert an exemplary sensor trigger  \texttt{"M007 ON"} to \textit{"motion sensor} \texttt{in bedroom fired with value} \textit{ON"}.
This description removes any sensor-identifying information such as sensor-id and only includes the universal semantics such as its symbolic location in the house and the sensor's binary state (ON/OFF).

The sensor type defines the type of the triggered sensor--motion, temperature, door--and its symbolic location, i.e., which room it is installed in, is manually encoded based on the home layout.

Sensor values are added to the sentence verbatim except when it is an integer or floating point number in which case we first convert it to text. 
For example, the sensor value of 22 will be converted to "twenty-two" whereas "ON" will be added to the sentence without any change. 
Complete specifications and examples for this coding scheme are given in Appendix \ref{appendix_tdost_basic} .

\subsubsection{TDOST-Temporal:} 

For this variant, we abstract from absolute timestamps of sensor triggers. 
Instead, we hypothesize that capturing the relative time gap between two consecutive sensor triggers could lead to better representations of the underlying activities \cite{bettadapura2013augmenting}. 
As such, in TDOST-Temporal,  we add said time difference between consecutive triggers to the second of the two sensor triggers involved.
We also add more context information to the location part such that, for example, 
``bedroom'' as the location for a sensor trigger is represented as ``bedroom near the front door''.
This variant also follows a template-based approach for which we use either of the two following templates  for the sensor triggers in a sequence:

\begin{quote}
    1: \texttt{<Temporal information> <Sensor-Type> sensor in <Location of Sensor> fired with value <Sensor-Value> }
\end{quote}

\begin{quote}
2: \texttt{<Sensor-Type> sensor in <Location of Sensor> fired with value <Sensor-Value> <Temporal information> }
\end{quote}

With these, we convert an exemplary  sensor trigger (of the sequence of two that are considered) \texttt{"T002 24"} to \texttt{"}\textit{five seconds later}\texttt{, the temperature sensor in kitchen} \textit{near stove}  \texttt{fired with value twenty-four".}
In this case, the \textit{five seconds later} is the time difference between the current sensor reading and the previous one. 

However, if the sensor trigger is the first in the activity sequence, the absolute time information is added at the end of the sentence. 
So, an exemplary \texttt{"T002 24"} is converted to  \texttt{"the temperature sensor in kitchen} \textit{near stove} \texttt{fired with value twenty-four} \textit{at seven hours thirty minutes AM}\texttt{"}.
Complete specifications and examples for this coding scheme are given in Appendix \ref{appendix_tdost_temporal}

\subsubsection{TDOST-LLM:} \label{tdostv3}
In the previous versions of TDOST, we manually converted the sensor data to text sentences. 
However, converting sensor triggers to text sentences provides an opportunity to augment the existing smart home datasets by automatically generating diverse sentences for each sensor reading.
For TDOST-LLM we utilize state-of-the-art large language models (specifically GPT-4.0) to generate diverse textual descriptions for each sensor reading.  \cite{achiam2023gpt}.
We provide the following prompt to GPT-4.0:

\begin{quote}
\texttt{
"You are an AI assistant that is helping in generating diverse texts and 
adding context to each sensor reading leveraging world knowledge.
Please generate diverse text sentences (3) for each sensor trigger.
You will be given a window of 5 sensor triggers. The given sensor trigger has the format: (Day of Week, Time on that day, sensor type, location context of the sensor,  Value of the Sensor).
The output should be a JSON (key : (Day of Week, Time on that day, sensor type, location context of the sensor,  Value of the Sensor) ) containing a list of the generated sentences. 
Sensor Trigger Window: }

\vspace*{0.5em}
\texttt{('Monday', 'Early Morning', 'Motion', 'between aisle and bathroom', 'OFF')}

\vspace*{0.5em}
\texttt{('Monday', 'Early Morning', 'Motion', 'bathroom', 'OFF')}

\vspace*{0.5em}
\texttt{('Monday', 'Early Morning', 'Motion', 'second bedroom', 'ON')}

\vspace*{0.5em}
\texttt{('Monday', 'Early Morning', 'Motion', 'second bedroom', 'OFF')}

\vspace*{0.5em}
\texttt{('Monday', 'Early Morning', 'Motion', 'between aisle and second bedroom', 'ON')"}
\end{quote}

Sensor readings are sent as input to an LLM in the form of a tuple combining the day of the week, time of the day, sensor type, location context of the sensor, and its value.
The sensor type, the sensor value, and the location context are the same as previously defined for TDOST-Temporal.
To limit the number of prompts to the LLM, we convert the absolute time given in each sensor trigger to higher granularity such as the day of the week and the relative period of the day. 
We use six different relative periods of the day namely, \textit{"Night"} (hours 0 to 5), \textit{ "Early Morning"} (hours 5 to 8), \textit{"Morning"} (hours 8 to 12), \textit{"Afternoon"} ( hours 12 to 17), \textit{"Evening"} (hours 17 to 21) and \textit{"Late Night"} (hours 21 to 23). 

For example, the sensor trigger \texttt{(2009-02-26 07:42:11.666589, M38, OFF)}  is first re-formatted to  \texttt{('Thursday', 'Early Morning', 'Motion', 'bathroom', 'OFF')} and then added to prompt for which GPT generates three different text sentences as given below:
\begin{quote}
\texttt{
        "As the first light of Thursday's early morning peeked through the windows, 
        the motion sensor in the bathroom remained inactive, suggesting that the 
        household was still embraced by sleep."
        }

\vspace*{0.5em}
\texttt{
        "Despite it being early on a Thursday morning, there was no movement detected 
        in the bathroom; the motion sensor's `OFF' status indicated a quiet start to 
        the day."
    }
    
\vspace*{0.5em}
\texttt{
        "The stillness of the house on this Thursday early morning was confirmed by 
        the motion sensor in the bathroom, which showed no signs of activity with its 
        `OFF' reading."
        }
\end{quote}
Complete specifications and examples for this coding scheme are given in Appendix \ref{appendix_tdost_llm}

\subsubsection{TDOST-LLM+Temporal:}  
For this encoding variant, we follow the same process of generating three different sentences using LLM. 
However, as an additional post-processing step, we also add the relative time difference to each of the GPT-generated sentences.
The relative time difference helps in adding more fine-granular temporal information to the sentences and is the same as defined in TDOST-LLM. 
We follow either of the following templates for TDOST-LLM+Temporal generation for the sensor triggers of a sequence:

\begin{quote}
    1: \texttt{<GPT-generated Sentence1> <Temporal information> , <GPT-generated Sentence2> <Temporal information> , <GPT-generated Sentence3> <Temporal information>}

    \vspace*{0.5em}
    2: \texttt{<Temporal information> <GPT-generated Sentence1> , <Temporal information> <GPT-generated Sentence2>, <Temporal information> <GPT-generated Sentence3> }
\end{quote}    
This more fine-granular temporal information is in addition to the day of the week and time of day added in TDOST-LLM. 
In TDOST-LLM+Temporal, we add a time difference between sensor triggers to the start of each sentence generated by LLM. 
Only in cases, where the sentences are generated for the first reading of activity sequence,  absolute time in text format is added at the end of the sentence. 
The process is the same as the one described previously for adding temporal information in TDOST-Temporal.

In the example given for TDOST-LLM in Section \ref{tdostv3}, we add the relative time difference as: 

\begin{quote}
\texttt{
 "After seven seconds, as the first light of Thursday's early morning peeked through the windows, 
the motion sensor in the bathroom remained inactive, suggesting that the 
household was still embraced by sleep."
}

\vspace*{0.5em}
\texttt{
"After seven seconds, despite it being early on a Thursday morning, there was no movement detected 
in the bathroom; the motion sensor's `OFF' status indicated a quiet start to 
the day."
}

\vspace*{0.5em}
\texttt{
"After seven seconds, the stillness of the house on this Thursday early morning was confirmed by 
the motion sensor in the bathroom, which showed no signs of activity with its 
`OFF' reading."
}
\end{quote}
Complete specifications and examples for this coding scheme are given in Appendix \ref{appendix_tdost_llm_temp}

\subsection{Layout Agnostic Human Activity Recognition using TDOST:}\label{sup-har}

The training process for deriving activity recognition models for source smart homes essentially follows the established, state-of-the-art supervised training approaches (e.g., \cite{liciotti2020sequential}). 
However, our TDOST-based HAR model learns from the novel, contextualized text descriptions and exploits the rich representation capability of language models. 
Once the HAR model has been trained on TDOST using labeled data from a source home, it can be used for inference of a common set of activities in different, target smart homes -- thereby not requiring any training nor adaptation for the target homes other than simply converting the sensor readings using the template-based approaches mentioned above that result in TDOST representations for the target home(s). 

Overall the application scenario of TDOST-based HAR in smart homes resembles a two-stage process:
\begin{description}
    \item [1. Training the HAR model on a Source Home using TDOST:]
    To perform end-to-end supervised training in the source home, we assume access to labeled source training data 
    The source home training process is illustrated in the first half of Fig.\ \ref{fig:method} and is as follows: 
    \begin{enumerate}
        \item First, raw sensor data from the labeled source dataset is converted to textual descriptions using one of the four variants described in Section \ref{TDOST-Variants}.
        
        \item Once the text sentences have been constructed, we deploy a frozen pre-trained language model,  specifically, yet without being limited to it, Sentence Transformers \cite{reimers2019sentence} to learn rich semantic text representations.
        
        \item Further, a feature encoder (Bi-LSTM \cite{liciotti2020sequential} unless specified otherwise in Section \ref{sec:results}) is deployed to capture the long-term dependencies from the sequence of textual representations.
        
        \item Finally, a classifier head is trained to map the feature representations to  activity labels.
    \end{enumerate}

    \item [2. Deploying the trained HAR model to Target Homes:]
    To deploy HAR models trained as outlined above in a different smart home, we collect the floorplan with information about sensor arrangements and modalities from the target home. 
    As a data pre-processing step, this meta-data is then used to create the templates that automatically map the target sensor triggers to text sentences as explained above. 
    Recognizing sets of standard activities in the target home using the unaltered source models is illustrated in the second half of Figure \ref{fig:method} and is as follows:
  \begin{enumerate}
        \item Similar to the source smart home, the raw test sensor data from the target home are converted to textual descriptions utilizing the target floorplan and sensor information along with the chosen template as explained above.
        
        \item The target textual descriptions are then converted into representations using the same pre-trained Sentence Transformer that was initially used to generate the source TDOST embeddings.  
        
        \item Subsequently, the target text embeddings are inputted into the frozen HAR model originally trained on source data, to predict activities in the target environment.  
    \end{enumerate}
\end{description}
In our approach, we assume that the activities covered by the HAR system are the same or at least similar in both source and target houses, allowing direct deployment of the source classifier head in the target homes. 
The main contribution of our approach lies in the facilitation of layout agnostic recognition across smart homes relying solely on readily available meta-information such as the target house's floorplan and coarse sensor specifications.
Furthermore, we keep the weights of the sentence encoder, feature encoder, and classifier head  (trained using source data) frozen when inferring on target datasets, demonstrating that no re-training or model adaptation occurs in the target homes.

\section{Setup}
\label{sec:settings}
In the previous sections we motivated our overall approach to layout-agnostic HAR in smart homes through our novel TDOST representations.
In what follows we now provide details of our experimental evaluation study that aims at assessing the effectiveness of our approach in realistic deployment settings.
In particular we focus on: 
\textit{(i)} Description of datasets;
\textit{(ii)} Data pre-processing; and
\textit{(iii)} Model architecture and Training Settings.

\subsection{Datasets}
We perform our experiments on four publicly available datasets from the Center for Advanced Studies in Adaptive Systems (CASAS), namely Aruba, Milan, Kyoto7, and Cairo \cite{cook2012casas}.
The details of these datasets, in terms of sensor modalities, residents, floorplan, and activities are presented in Table \ref{tab:datasets}: 
The Aruba smart home is the largest in terms of the number of data points (hence more training data), while being balanced in terms of sensor modalities ($M, D,$ and $T$).
Its floorplan is similar to the next largest dataset, Milan, which also include Motion ($M$), Door ($D$), and Temperature ($T$) sensors.

\begin{table*}[t]
	\centering
	\small
	\begin{tabular}{P{0.7cm} P{1.5cm} P{1cm} P{4cm}  P{7cm}}
        \toprule
		Dataset & Sensors & Residents & Floorplan & Activities (\# Data Points) \\ 
		\midrule
		Aruba & $[M, D, T]$ & 1 & Single story home with living space, dining space, kitchen, office, 2 bedrooms, 2 bathrooms, and closet & Relax (2919), Meal Preparation (1606), Enter Home (431), Leave Home (431), Sleeping (401), Eating (257), Work (171), Bed to Toilet (157), Wash Dishes (65), Housekeeping (33), Resperate (6), Other (6354) \\ \hline
            Milan & $[M, D, T]$ & 1 & Single story home with living space, dining space, kitchen, workspace/TV room, 2 bedrooms, 2 bathrooms, and closet & Kitchen Activity (554), Guest Bathroom (330), Read (314), Master Bathroom (306), Leave Home (214), Master Bedroom Activity (117), Watch TV (114), Sleep (96), Bed to Toilet (89), Desk Activity (54), Morning Meds (41), Chores (23), Dining Room Activity (22), Evening Meds (19), Meditate (17), Other (1943)  \\ 
            \hline
            Kyoto7 & $[M, D, T]$\newline$[I, LS, AD]$ & 2 & Double story home with living space, dining space, kitchen with pantry and closet, 2 bedrooms, office, bathroom, and closet  & Meal Preparation (107), R1 Work (59), R1 Personal Hygiene (44), R2 Work (44), R2 Bed to Toilet (39), R2 Personal Hygiene (38), R1 Sleep (35), R2 Sleep (35), R1 Bed to Toilet (34), Watch TV (30), Study (9), Clean (2), Wash Bathtub (1)  \\ \hline
            Cairo & $[M, T]$ & 2 & Three story home with living space, dining space, kitchen, 2 bedrooms, office, laundry room, and garage room & Leave Home (69), Night Wandering (67), R1 Wake (53), R2 Wake (52), R2 Sleep (52), R1 Sleep (50), Breakfast (48), R1 Work in Office (46), R2 Take Medicine (44), Dinner (42), Lunch (37), Bed to Toilet (30), Laundry (10) \\ 
		%
		
		\bottomrule
	\end{tabular}
 \caption{Overview of the datasets used in our experiments. 
 The sensors $M$, $D$, $T$, $I$, $LS$, and $AD$ represent motion, door, temperature, item, light switch, and activate device (burner, hot water, and cold water) sensors. 
 The selected four datasets vary in terms of sensor modalities, sensor arrangements, number of residents, floorplans, and activities. 
 }
	\label{tab:datasets}
\end{table*}

Kyoto7 and Cairo smart homes are multi-story setups with distinct locations like the kitchen pantry, laundry room, garage, etc. 
They also have a comparatively smaller number of data points than Aruba and Milan, while having two residents (hence different idiosyncrasies in performing activities). 
However, Cairo and Kyoto7 smart homes differ drastically in terms of their sensor modalities, where Cairo has only $M$ and $T$ sensors, whereas, Kyoto7 has all $M$, $D$, $T$, Item (use) $I$, Light switch $LS$, and Activate Device $AD$ sensors.  

In terms of activities, for all homes both common (such as sleep and Bed to Toilet) and uncommon (such as Resperate, Meditate, and Wash Bathtub) activities have been collected and annotated.
Since we do not assume access to any data from a target  home, we convert the activities into a common set: $\{$Relax, Cook, Leave Home, Enter Home, Sleep, Eat, Work, Bed to Toilet, Bathing, Take Medicine, Personal Hygiene, and Other$\}$, which is in line with related work in the field \cite{liciotti2020sequential}. 
The details of this mapping of activities is detailed in  Appendix \ref{sec:appendix:activity_mapping}.  
While performing the standard HAR experiment as baselines (Section \ref{subsec:results_standard_har}), we consider these common activities as our class labels. 
For our layout-agnostic HAR experiments, i.e., the main target of our approach, (Section \ref{subsec:results_layout-agnostic_har}), we first take the intersection of the common activity sets from source and target smart homes. 
A HAR model is trained on source data only and w.r.t.\ the aforementioned set of standard activities defined through the  intersection as described above.
See Section \ref{sec:discussion:limitations} for a discussion on the--minimal--practical implications of this design choice.

\subsection{Data Preprocessing}
We convert sequences of sensor data into windows of 100 sensor triggers per activity class for each dataset.
Each of these windows of sensor triggers is passed through the sentence encoders to generate a sequence of textual embeddings. 
Once the the textual embeddings have been generated, we shuffle the activity windows and produce three stratified folds ensuring that the proportion of activity labels across folds is identical. 
Further, 20\% of windows are randomly selected from the train split of every fold to form a validation split.

For experiments using TDOST-LLM and TDOST-LLM+Temporal variants,  the three LLM-generated sentences for every sensor trigger are used to construct three different data points. 
This implies that for every activity window in the original dataset we produce three different TDOST windows (using the LLM) which are further shuffled to be split into 3 folds and train-validation-test split. 
We report averages and standard deviations of test accuracy and weighted F1-score (used as F1-score throughout the paper) across three folds, which is in line with previous work \cite{liciotti2020sequential}.

\subsection{Model Architecture and Training Settings}
\label{subsec:model_architecture}
In our experiments we employ the state-of-the-art, pre-trained Sentence Transformers library \cite{reimers2019sentence} for generating embeddings for text sentences describing the sensor triggers.
Specifically,  we use the ``all-distilroberta-v1'' model variant of Sentence Transformers which maps the input sentence to a 768-dimensional dense vector space. 
In this model, the pre-trained DistilRoBERTa \cite{Sanh2019DistilBERTAD, liu2019roberta} is fine-tuned with 1B sentence pairs using a contrastive learning objective that forces the model to predict which sentence out of a set of randomly sampled sentences should be paired with current sentence.
The underlying DistillRoberta model is trained on the OpenWebText Corpus \cite{Gokaslan2019OpenWeb}.

The sentence embeddings for a sequence of sensor triggers are then forwarded to a Bi-LSTM backbone, a feature encoder that captures long-range dependencies.
The Bi-LSTM model employs a bidirectional LSTM architecture that utilizes forward and backward passes for output prediction.
It is similar to the one proposed for DeepCASAS \cite{liciotti2020sequential} and uses 64 hidden units.
These encoder representations are finally mapped to the probability distribution of output activity classes using a classification head of one linear layer.

Model training on source data is supervised  using Cross-Entropy loss with access to labeled data from the source home.
All models have been implemented in Pytorch \cite{paszke2019pytorch}. 
For HAR model training, we perform a grid search over different learning rates ([1e-3, 1e-4, 5e-4]) and weight decay ( [0.0, 1e-4, 1e-5]).  
When training the DeepCASAS model, we use a learning rate of 0.001 and no weight decay as specified in the previous related work \cite{liciotti2020sequential}. 
We train all models for 75 epochs and use Adam\cite{kingma2014adam} optimizer for optimizing the model weights.

\section{Results}
\label{sec:results}

In this section, we present and discuss the results of our experimental evaluations primarily focusing on various source-target dataset combinations.
We analyze the performance of our approach considering the diversity of source-target transfer conditions as well as the TDOST-variants used to create text sentences. We also discuss the impact of using different encoders such as diverse Sentence Transformer variants and feature encoders, on downstream activity recognition performance in target homes.
The results of our approach were compared against the state-of-the-art supervised baseline, DeepCASAS \cite{liciotti2020sequential}.

\subsection{Layout-Agnostic Human Activity Recognition using TDOST}
\label{subsec:results_layout-agnostic_har}
TDOST effectively facilitates the training of a HAR classifier using only labeled trained data from a  source home -- that subsequently can be used without modification at target houses with different floorplans and sensor installations.  
In our first set of experiments, we test the ability of TDOST to perform such layout-agnostic HAR thereby targeting a range of smart homes as listed in Table \ref{tab:datasets}. 
We analyze our approach using DeepCASAS \cite{liciotti2020sequential} as the classification backend and present the results in Table \ref{tab:layout_agnostic_tdost}. 
Based on our empirical evaluation, we can draw the following insights in performing layout-agnostic HAR.

\begin{table}[]
\centering
\resizebox{\textwidth}{!}{
\begin{tabular}{ c cc cc cc }
\toprule
\multirow{2}{*}{Method (Source Aruba)} & \multicolumn{2}{c}{Milan} & \multicolumn{2}{c}{Cairo}  & \multicolumn{2}{c}{Kyoto7} \\
                        & Acc        & F1           & Acc        & F1        & Acc        & F1        \\ \hline
\textit{TDOST-Basic}    & $67.57 \pm 0.52$  & $59.79 \pm 0.90$ & $65.73 \pm 0.82$ & $53.09 \pm 1.05$ & $42.01 \pm 0.25$ & $28.90 \pm 0.56$ \\

\textit{TDOST-Temporal}  &  $69.08 \pm 0.76$ & $61.93 \pm 0.20$ & \cellcolor[HTML]{caebc0} $\textbf{66.35} \pm \textbf{0.64}$ &$53.09 \pm 0.82$ & $45.83 \pm 1.47$ & $33.45 \pm 1.29$ \\

\textit{TDOST-LLM}   & $73.57 \pm 1.16$ &$70.60 \pm 1.38$ & $60.45 \pm 0.94$ & $54.00 \pm 0.91$ & \cellcolor[HTML]{caebc0} $\textbf{51.91} \pm \textbf{2.41}$ & \cellcolor[HTML]{caebc0} $\textbf{42.79} \pm \textbf{1.91}$\\

\textit{TDOST-LLM+Temporal}  & \cellcolor[HTML]{caebc0}$\textbf{75.78} \pm \textbf{0.70}$ & \cellcolor[HTML]{caebc0} $\textbf{73.10} \pm \textbf{0.90}$ & $63.48 \pm 0.98$  & \cellcolor[HTML]{caebc0} $\textbf{55.15} \pm \textbf{0.57}$ & $48.32 \pm 2.39$ & $40.73 \pm 2.33$ \\

\textit{DeepCASAS}    & $50.47 \pm 0.75$ & $44.63 \pm 0.09$  & $50.94 \pm 3.98$   &$48.50 \pm 2.82$  & $34.55 \pm 0.25$ & $20.45 \pm 0.61$ \\

\midrule 

\multirow{2}{*}{Method (Source Milan)} & \multicolumn{2}{c}{Aruba} & \multicolumn{2}{c}{Cairo} & \multicolumn{2}{c}{Kyoto7}\\
                        & Acc        & F1           & Acc        & F1        & Acc        & F1      \\ \midrule
\textit{TDOST-Basic}   & $63.79 \pm 0.08$  & $52.24 \pm 0.09$ & $49.13 \pm 2.98$  & $45.82 \pm 2.52$ & $38.72 \pm 0.49$ & $28.96 \pm 1.29$   \\

\textit{TDOST-Temporal}   &$63.51 \pm 0.45$  &$53.33 \pm 0.32$  &$45.92 \pm 3.32$ &  $41.55 \pm 2.40$ & $40.62 \pm 2.79$ & $39.11 \pm 2.49$ \\

\textit{TDOST-LLM}   & $73.62 \pm 0.08$ &  $72.46 \pm 0.10$ & $49.94 \pm 1.39$  & $45.48 \pm 1.18$ & \cellcolor[HTML]{caebc0} $\textbf{51.97} \pm \textbf{2.15}$ & \cellcolor[HTML]{caebc0} $\textbf{47.27} \pm \textbf{1.95}$ \\

\textit{TDOST-LLM+Temporal}   & \cellcolor[HTML]{caebc0}$\textbf{77.23} \pm \textbf{0.48}$  & \cellcolor[HTML]{caebc0} $\textbf{76.23} \pm \textbf{0.45}$  & \cellcolor[HTML]{caebc0} $\textbf{57.93} \pm \textbf{0.22}$  & \cellcolor[HTML]{caebc0} $\textbf{51.90} \pm \textbf{0.32}$ & $51.16 \pm 1.78$ & $46.38 \pm 1.74$ \\

\textit{DeepCASAS}    & $37.38 \pm 0.69$ &  $28.61 \pm 0.45$  &  $34.55 \pm 2.15$  & $35.72 \pm 1.27$ &  $39.24 \pm 0.89$ & $28.44 \pm 0.78$\\

\midrule

\multirow{2}{*}{Method (Source Cairo)} & \multicolumn{2}{c}{Milan} & \multicolumn{2}{c}{Aruba} & \multicolumn{2}{c}{Kyoto7} \\
                        & Acc        & F1           & Acc        & F1           & Acc        & F1    \\ \midrule
\textit{TDOST-Basic}   &$60.34 \pm 1.09$  & $64.39 \pm 0.91$ & \cellcolor[HTML]{caebc0} $\textbf{77.90} \pm \textbf{0.51}$  & $71.47 \pm 0.32$ & \cellcolor[HTML]{caebc0} $47.31 \pm 1.52$ & $32.71 \pm 2.35$ \\

\textit{TDOST-Temporal}   &$67.41 \pm 0.81$ &$67.45 \pm 0.47$ &$77.88 \pm 0.02$  &  $71.54 \pm 0.12$ & $44.62 \pm 0.76$ & $32.57 \pm 1.00$  \\

\textit{TDOST-LLM}   & $63.49 \pm 0.18$ & $65.54 \pm 0.27$ & $64.28 \pm 0.63$ &$66.82 \pm 0.43$ &  $44.12 \pm 1.11$ & \cellcolor[HTML]{caebc0}  $33.82 \pm 1.35$   \\

\textit{TDOST-LLM+Temporal}  & \cellcolor[HTML]{caebc0} $67.99 \pm 0.68$ & \cellcolor[HTML]{caebc0}  $\textbf{69.30} \pm \textbf{0.59}$ & $74.25 \pm 0.54$ & \cellcolor[HTML]{caebc0} $\textbf{71.97} \pm \textbf{0.40}$ & $43.90 \pm 0.28$ &  $31.64 \pm 0.53$\\

\textit{DeepCASAS}     & $\textbf{68.94} \pm \textbf{0.41}$  & $64.27 \pm 0.40$ &  $65.85 \pm 0.76$  & $67.01 \pm 0.72$ & $\textbf{53.83} \pm \textbf{1.93}$ &  $\textbf{43.43} \pm \textbf{2.48}$ \\

\midrule

\multirow{2}{*}{Method (Source Kyoto7)} & \multicolumn{2}{c}{Aruba} & \multicolumn{2}{c}{Cairo} & \multicolumn{2}{c}{Milan} \\
                        & Acc        & F1           & Acc        & F1         & Acc        & F1     \\ \midrule
\textit{TDOST-Basic}  & $40.58 \pm 0.62$ & $37.54 \pm 0.70$ & $66.34 \pm 2.70$ & $67.24 \pm 1.95$ & $56.46 \pm 1.49$ & $56.37 \pm 1.41$ \\

\textit{TDOST-Temporal}   & $49.62 \pm 3.93$ & $47.71 \pm 4.60$ & $55.16 \pm 1.73$ & $61.88 \pm 1.41$ & $56.74 \pm 1.80$ & $57.76 \pm 1.29$   \\

\textit{TDOST-LLM} & \cellcolor[HTML]{caebc0} $\textbf{50.83} \pm 1.20$ & \cellcolor[HTML]{caebc0} $\textbf{51.07} \pm 1.23$ & \cellcolor[HTML]{caebc0} $\textbf{73.59} \pm 0.59$ & \cellcolor[HTML]{caebc0} $\textbf{73.59} \pm 0.61$ &  $57.22 \pm 0.50$ &  $59.08 \pm 0.54$ \\

\textit{TDOST-LLM+Temporal}  & $49.41 \pm 1.64$ & $49.44 \pm 1.71$ & $67.78 \pm 3.78$ & $70.52 \pm 2.47$ & \cellcolor[HTML]{caebc0} $59.79 \pm 0.23$ & \cellcolor[HTML]{caebc0} $60.84 \pm 0.22$ \\

\textit{DeepCASAS}    & $39.62 \pm 1.82$ &  $41.64 \pm 2.30$ & $49.18 \pm 7.36$ &  $54.38 \pm 5.05$ & $\textbf{62.59} \pm 0.65$ & $\textbf{62.50} \pm 0.55$ \\

\bottomrule

\end{tabular}
}
\caption{Performance of the TDOST versions in the layout-agnostic HAR setup. 
The overall best results in every configuration are boldfaced, whereas, the best-performing TDOST version is highlighted in  green. 
Overall, TDOST-based layout-agnostic HAR significantly outperforms the DeepCASAS baseline demonstrating the effectiveness of our approach.
We also observe that incorporating temporal information, additional context, and augmentations to  TDOST (the four variants) leads to substantial performance improvements.
}
\vspace*{-1em}
\label{tab:layout_agnostic_tdost}
\end{table}

\subsubsection{Performance Improvement in Layout-Agnostic HAR using TDOST}
Out of a total of 12 ($4 \times 3$) source-target settings, our TDOST versions outperform DeepCASAS in 10 of these settings, with an overall improvement of $\textbf{18.76\%}$ and $\textbf{17.87\%}$ in accuracy and F1 score, respectively. 
Moreover, in settings like Aruba$\rightarrow$Milan, Milan$\rightarrow$Aruba, Cairo$\rightarrow$Aruba, and Kyoto7$\rightarrow$Cairo, we are able to achieve $\textbf{70\%+}$ accuracy and F1 score, without using a single data point from the target smart home during training. 
These results testify to our claim that TDOST is able to learn a layout-agnostic HAR model that can be transferred across diverse smart homes. 

\subsubsection{Addressing Diverse Source-Target Conditions}
Analyzing the HAR performances across homes (Table \ref{tab:layout_agnostic_tdost}), it is important to keep in mind the transfer conditions between the source$\rightarrow$target settings. 
In particular, we refer to the sensor modalities present in each smart home, the number of residents, floorplans, activities, and the activity translations (Tables \ref{tab:datasets} and  \ref{tab:activity_translation}) that differ for the various source$\rightarrow$target settings
For example, the transfer of the layout-agnostic HAR model between Aruba and the Milan smart homes (Aruba$\rightarrow$Milan and Milan$\rightarrow$Aruba) is relatively easier than the same with Cairo or Kyoto7 smart homes. 
This is because both Aruba and Milan have the same sensor modalities, number of residents, and similar floorplans (both are single-story homes with the same number of bedrooms, bathrooms, offices, etc.). 

Moreover, both Aruba and Milan homes come with larger amounts of sample data.
Despite such seemingly favorable conditions, the model trained without TDOST fails to transfer: In the Aruba$\rightarrow$Milan setting it can reach only around $50\%$ accuracy and $44\%$ F1 score.
The situation is even worse for the Milan$\rightarrow$Aruba setting with only around $37\%$ accuracy and $28\%$ F1 score.
In contrast, our TDOST-based models outperform these state-of-the-art models by staggering $\textbf{32.58\%}$ and $\textbf{38.04\%}$ in accuracy and F1 score, respectively.

Whenever the source datasets are Aruba and Milan, irrespective of the target dataset, we see large performance improvements overall ($\textbf{22.32\%}$ accuracy, $\textbf{23.34\%}$ F1) when using TDOST (versions). 
Remarkably, for settings like Aruba$\rightarrow$Kyoto7, Aruba$\rightarrow$Cairo, Milan$\rightarrow$Kyoto7, and Milan$\rightarrow$Cairo, where the transfer conditions are less favorable (unseen sensor modalities of $I, LS,$ and $D$, different floorplans, and different number of residents), the performance increase is still substantial, demonstrating the overall effectiveness of our approach.

Finally, in the least favorable transfer conditions, i.e., when the source homes are Cairo and Kyoto7 (with under/over-representation of sensor modalities, differences in floorplans, number of residents, and most importantly, smaller sizes of available training data), TDOST offers variable levels of performance improvements. 
For Kyoto7$\rightarrow$Aruba or Kyoto7$\rightarrow$Cairo, we see substantial performance improvements ($17.81\%$ and $14.32\%$ in accuracy and F1 score, respectively), whereas for Cairo$\rightarrow$Aruba and Cairo$\rightarrow$Milan, we only see marginal improvements ($5.55\%$ and $4.99\%$ in accuracy and F1 score, respectively). 
In the two exceptional cases (Cairo$\rightarrow$Kyoto7 and Kyoto7$\rightarrow$Milan), we observe a marginal performance drop of $4.66\%$ and $5.63\%$ in accuracy and F1 scores, respectively, after using TDOST.
We discuss these two exceptional cases in more detail in Section \ref{subsubsec:exxception}.
Overall, we conclude that a source dataset should have more training data samples for training TDOST-based HAR system, rather than more variety of sensor modalities or similarity in floorplans and other aspects.  
See Sec. \ref{sec:discussion:transfer-knowledge} where we discuss ways to address this.  

\subsubsection{Analysis of TDOST Versions}
While designing the TDOST versions, we incrementally developed them to include more context, temporal information, and augmentations, with the goal of testing which ones of these factors play a crucial role in the layout-agnostic HAR setup.
Out of the 12 source-target settings, the best performing TDOST versions in 7 of them use the additional context and temporal information (the \textit{-Temporal} versions).
On the other hand, in all 12 of the source-target settings, the best-performing TDOST version uses LLMs to generate augmentations of the textual descriptions.   
We see a marginal improvement of $0.57\%$ and $1.9\%$ in accuracy and F1 score when the temporal information is incorporated into TDOST, i.e., performance improvements from \textit{TDOST-Basic} to \textit{TDOST-Temporal}. 
However, when we add augmentations through the use of LLMs, i.e., from \textit{TDOST-Basic} to \textit{TDOST-LLM}, we observe a larger performance increase of $3.25\%$ and $7\%$ in accuracy and F1 scores, respectively. 
Based on these improvements, we recommend using \textbf{LLM-based augmentations as well as relative temporal context}  in TDOST to achieve effective activity recognition. 

\subsubsection{Exceptions and Ways to Improvements}
\label{subsubsec:exxception}
In settings Cairo$\rightarrow$Kyoto7 and Kyoto7$\rightarrow$Milan, we observe that TDOST marginally underperforms compared to the DeepCASAS setup. 
Before investigating the potential causes behind this, we point out that optimizing the choice of source smart home can potentially avoid such anomalous performances.
Specifically, smart homes like Cairo and Kyoto7 have multi-story floorplans, more than one resident, fewer data samples for training, uncommon activities, etc.
Without limiting the general applicability of our overall approach it is seemingly more effective to train on scenarios with more common floorplans, single residents, more training data, and commonly occurring activities in larger numbers.

We first consider the Cairo$\rightarrow$Kyoto7 setup (Table \ref{tab:datasets}).   
Cairo covers 7 activities: Sleep, Work, Eat, Leave Home, Bed to Toilet, Take Medicine, and the Other. 
Out of these, only 4 (Sleep, Work, Bed to Toilet, and Other) are common to the Kyoto7 dataset. 
The distribution of these activities in terms of number of data points in the source (Cairo) is -- Other: 748, Sleep: 102, Work: 56, and Bed to Toilet: 30, which indicates a large class imbalance in addition to fewer training data, especially for the minority classes. 
Further, Cairo contains only $M$ and $T$ sensors, whereas, Kyoto7 contains all of the $M$, $D$, $T$, $I$, $LS$, and $AD$ sensors.
Lastly, the majority class (Other) on the source dataset (Cairo) represents the noise as well as the ``Wake'' activity, which resembles the ``Bed to toilet'' activity in terms of location. 
In contrast, for the target dataset (Kyoto7), the Other class represents activities like ``Wash Bathtub'' and ``Study'', occurring at different locations in the smart home. 
Thus, we attribute the anomalous performance of TDOST in this source-target setting to the combined effect of a lack of (balanced) training data, large differences in the sensor modality sets, and a misrepresentation of activities in the majority class.
Similar transfer conditions can also be seen in the Kyoto7$\rightarrow$Milan setup.

A straightforward way to avoid such cases would be to select a better source dataset for training the TDOST-based HAR model. 
However, we focus on improving the performance of TDOST in \textit{any} given source-target transfer setting.
Specifically, we identify 3 major TDOST components that impact the overall effectiveness: 
(\textit{i}) construction of textual descriptions;
(\textit{ii}) choice of sentence transformer; and 
(\textit{iii}) choice of classifier. 
Our TDOST versions exploit different ways of constructing textual descriptions. 
Hence, in the next experiments, we further optimize the remaining two components, namely, the sentence transformers and the classifier.   

\subsection{Improving TDOST-based HAR: Choice of Sentence Transformer}
\label{subsec:ablation_st}
\begin{figure}[]
  \centering
  \includegraphics[width=\linewidth]{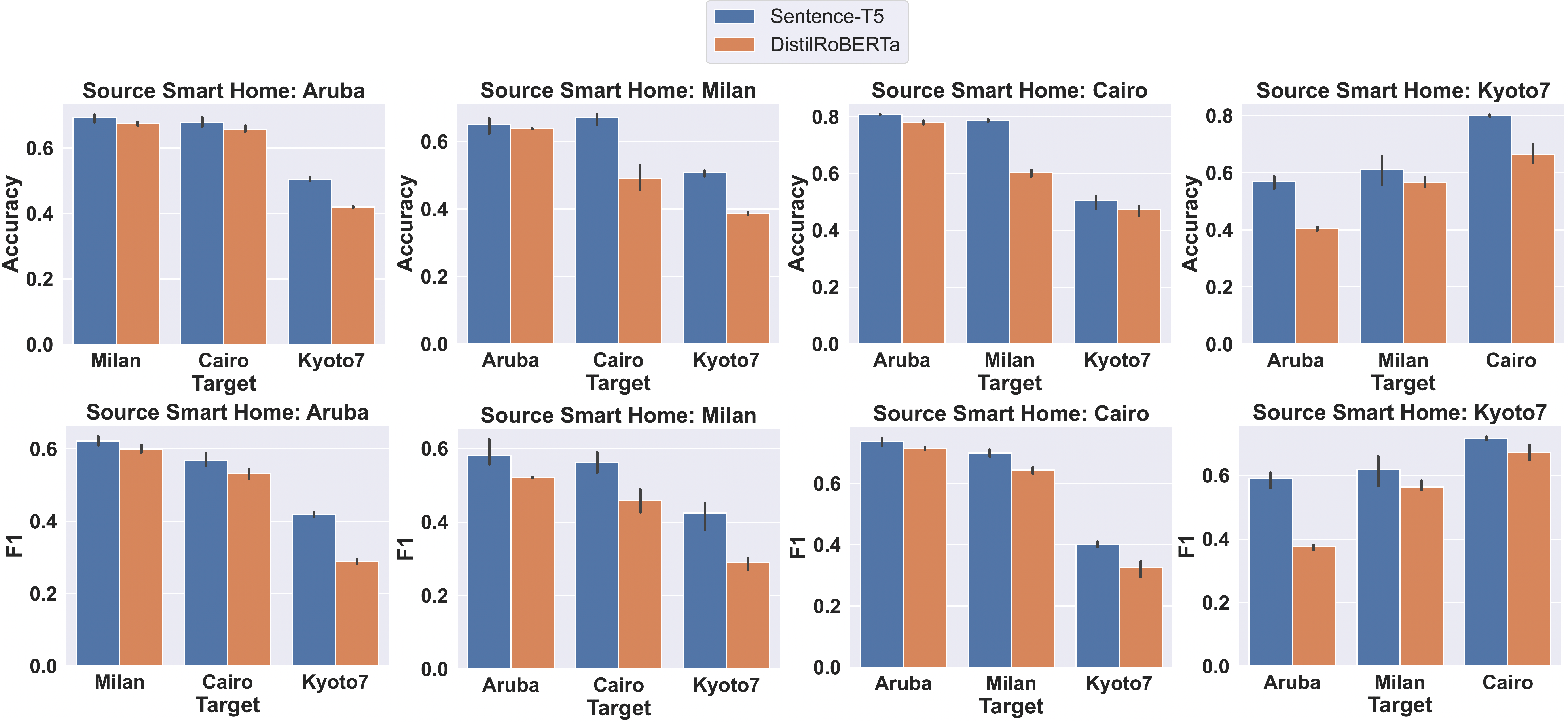} 
  \caption{Comparison between the usage of DistilRoBERTa and Sentence-T5 variants of the  Sentence Transformer models in the TDOST pipeline. 
  The Sentence-T5 variant, which is an encoder-decoder transformer, significantly outperforms Distill Roberta across all the source-target settings. 
  Moreover, the improvements are noticeable, especially in more challenging transfer settings involving the Cairo and Kyoto7 datasets.}

\label{fig:ablation_st}
\end{figure}

In our previous experiments, we used the  DistilRoBERTa variant of Sentence Transformer to encode the textual representations in the TDOST pipeline.
We now explore a different variant of Sentence Transformer, specifically the Sentence-T5 \cite{ni2021sentence} pre-trained model which uses an encoder-decoder transformer, unlike the encoder-only RoBERTa. 

In this experiment, we use the \textit{TDOST-Basic} version to generate textual descriptions and use a Bi-LSTM classifier described in Section \ref{subsec:model_architecture}. 
The remaining experimental setup is kept constant, where only the sentence transformers vary in the two settings. 
The results of this experiment are shown in Figure \ref{fig:ablation_st}. 
Across \textit{all} the source-target settings, we observe a performance improvement by replacing the DistilRoBERTa with Sentence-T5.
The overall improvement across all settings is $\textbf{8.58\%}$ and $\textbf{7.9\%}$ for accuracy and F1 score, respectively.
The noticeable improvements occur in more challenging settings like Kyoto7$\rightarrow$Aruba, Kyoto7$\rightarrow$Milan, Milan$\rightarrow$Kyoto7, Milan$\rightarrow$Cairo, and Aruba$\rightarrow$Kyoto7, serving our purpose of improving \textit{any} source-target transfer setting, irrespective of its inherent challenges (as mentioned at the end of Section \ref{subsubsec:exxception}).
We attribute this improvement to the encoder-decoder transformer structure of Sentence-T5, which ultimately produces better (more transferable) embeddings. 
Additionally, the Sentence-T5 variant has more model parameters (110 million) compared to  DistilRoBERTA (\textasciitilde82 Million), resulting in more representational power and better generalization.

\subsection{Improving TDOST-based HAR: Choice of Classifier}
\begin{figure}[]
  \centering
  \includegraphics[width=\linewidth]{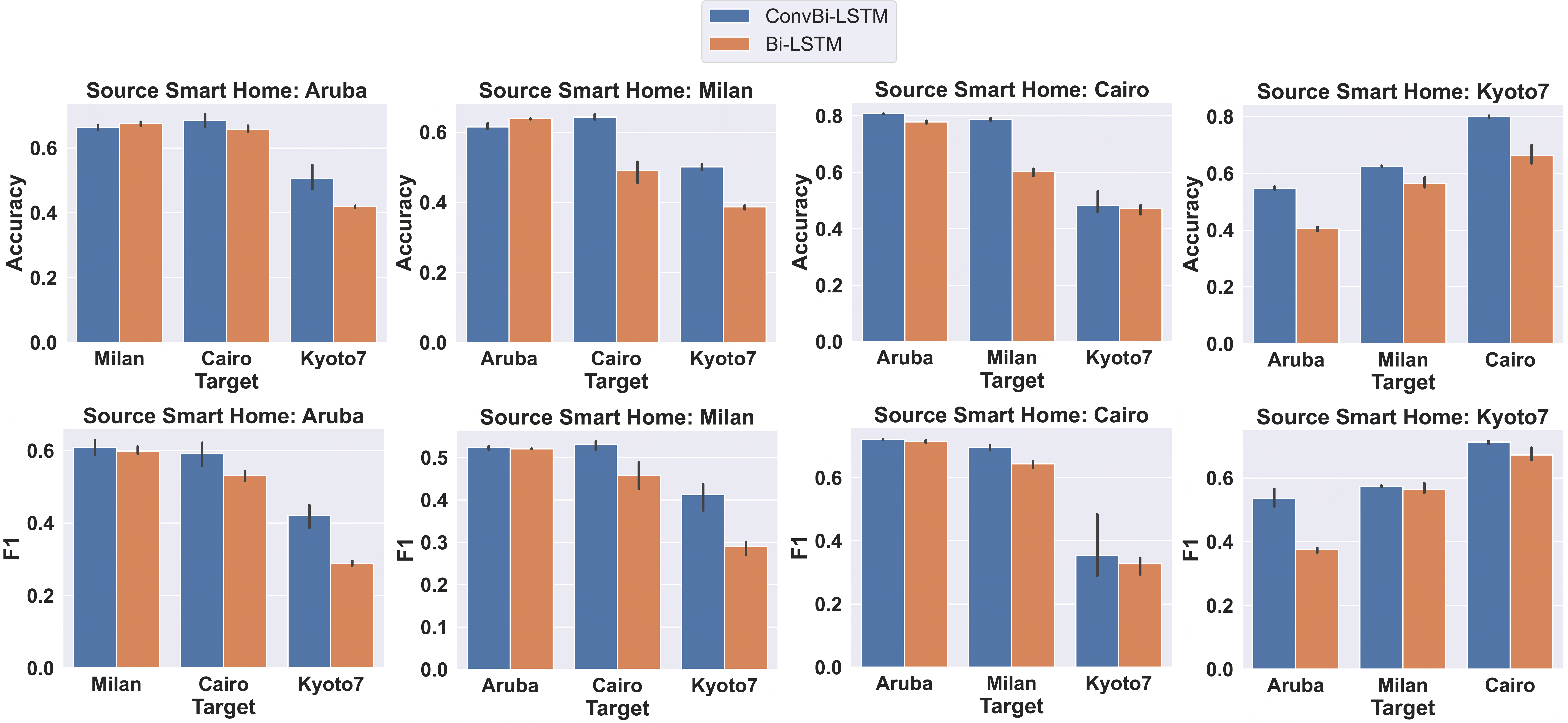} 
  \caption{Comparison between the usage of Bi-LSTM and ConvBi-LSTM classifiers in the TDOST pipeline. 
  Using a ConvBi-LSTM classifier, having more parameters in its architecture, significantly outperforms the Bi-LSTM classifier and provides performance improvements in almost all source-target transfer settings.
  These improvements are noticeable, especially in more challenging transfer settings involving the Cairo and Kyoto7 datasets.}

\label{fig:ablation_classifier}
\end{figure}

In this experiment, we focus on the choice of feature encoder in the classifier that is used in the source domain training (using TDOST embeddings). 
In our earlier experiments, we selected the Bi-LSTM encoder (described in Section \ref{subsec:model_architecture}) as it is one of the standard neural networks used in the relevant literature \cite{liciotti2020sequential}.
This choice was also inspired by our effort to make the comparisons fair with state-of-the art, in Table \ref{tab:layout_agnostic_tdost}
However, this architecture may not be optimal for our TDOST embeddings. 
This is because the number of (trainable) parameters in the Bi-LSTM architecture is set according to the dimensionality of the feature space of the data points (64 in the case of DeepCASAS), whereas, our embeddings are 768 dimensional. 
Hence, a high-capacity classifier is expected to work better when using TDOST embeddings.
In the HAR literature, ConvLSTM architectures are widely used, where they have performed well on sequential sensor data for HAR \cite{ordonez2016deep} 
Therefore, we use a ConvBi-LSTM architecture consisting of three 1-dimensional convolution layers followed by a recurrent Bi-LSTM layer. 

The results of this experiment are shown in Figure \ref{fig:ablation_classifier}. 
Except for two source-target settings (Aruba$\rightarrow$Milan and Milan$\rightarrow$Aruba), we observe a significant performance improvement, specifically of $7.49\%$ and $5.81\%$ for accuracy and F1 score, respectively. 
In the two settings the performance difference is marginal ($1.81\%$ accuracy drop and $0.64\%$ F1 gain).
Similar to our observation from Section \ref{subsec:ablation_st}, the performance improvements are eminent, especially in more challenging transfer scenarios (involving the Cairo and Kyoto7 datasets) like Aruba$\rightarrow$Kyoto7, Milan$\rightarrow$Cairo, Milan$\rightarrow$Kyoto7, Cairo$\rightarrow$Milan, Kyoto7$\rightarrow$Aruba, and Kyoto7$\rightarrow$Cairo.  
We attribute this improvement to the use of a higher-capacity classifier (more trainable parameters), especially given our transformed embedding space, from sensor triggers to TDOST embeddings.

\subsection{Human Activity Recognition using TDOST}
\label{subsec:results_standard_har}
In this final experiment, we test the effectiveness of TDOST as a choice for representing sensor data in a standard HAR setup, which refers to the case where both the classifier training and inference are performed on the data collected in the same smart home. 
This setting essentially serves as a baseline comparison for the general applicability of our TDOST-based modeling approach -- even for scenarios where transfer between homes is not targeted.
We compare our results with DeepCASAS \cite{liciotti2020sequential}, a widely recognized state-of-the-art HAR framework.  
Table \ref{tab:har_tdost} shows the results obtained for the four TDOST versions and DeepCASAS across the selected datasets.  
TDOST outperforms the DeepCASAS setup in three of the selected four datasets (Milan, Cairo, and Kyoto7), where the average gain in accuracy and F1 score is $\sim 2.296\%$ and $\sim 2.58\%$, respectively. 
On the Aruba dataset, TDOST performs comparably to DeepCASAS with a difference of only around $0.36\%$ in both accuracy and F1 score. 
These results show that the TDOST-generated representation space for sensor triggers is better (or equally good) for the supervised classification of activities in a smart home.  

Out of the proposed TDOST versions, \textit{TDOST-LLM+Temporal} performs the best in the case of three datasets (Milan, Cairo, and Aruba), whereas, the \textit{TDOST-Temporal} version performs the best for the Kyoto7 dataset. 
These results direct us to some takeaways that can be applied in constructing the textual descriptions for smart home sensor triggers.
Firstly, we observe that the result of incorporating temporal context (time difference between consecutive sensor readings) in textual descriptions significantly improves the HAR performance in the case of the Kyoto7 dataset. 
In the case of the other datasets, the performance increase/decrease as a result of this incorporation is not significant, showing that the aspect of temporal information is crucial for activity recognition in certain conditions only. 
Secondly, we observe that performing augmentations of textual descriptions using LLMs significantly improves the classification performance. 
Specifically, we see an overall increase of $2.7\%$ and $2.99\%$ in accuracy and F1 score, respectively, between \textit{TDOST-Basic} and \textit{TDOST-LLM} versions, whereas, the same between \textit{TDOST-Temporal} and \textit{TDOST-LLM+Temporal} is $2.02\%$ and $2.06\%$, respectively. 
Thus, based on our empirical evaluation, we recommend performing augmentations of textual descriptions to boost the HAR performance.

\begin{table}[]
\centering
\resizebox{\textwidth}{!}{
\begin{tabular}{c cc cc cc cc}
\toprule
\multirow{2}{*}{Method} & \multicolumn{2}{c}{Milan} & \multicolumn{2}{c}{Cairo} & \multicolumn{2}{c}{Aruba}   & \multicolumn{2}{c}{Kyoto7} \\
                        & Acc        & F1           & Acc        & F1           & Acc        & F1           & Acc        & F1  \\ \midrule
\textit{TDOST-Basic}   & $90.53 \pm 0.57$  & $90.32 \pm 0.38$  & $84.64 \pm 1.39$  & $84.75 \pm 1.00$  & $92.77 \pm 0.43$  & $92.62 \pm 0.44$   & $66.67 \pm 2.95$ & $65.29 \pm 4.21$\\

\textit{TDOST-Temporal}  & $89.32 \pm 0.70$  & $88.79 \pm 0.92$ &  $82.99 \pm 2.34$ &$83.17 \pm 2.07$  & $92.78 \pm 0.26$ & $92.64 \pm 0.21$ & \cellcolor[HTML]{caebc0}  $\textbf{74.31} \pm \textbf{2.89}$  & \cellcolor[HTML]{caebc0} $\textbf{74.19} \pm \textbf{2.86}$ \\

\textit{TDOST-LLM}  & $91.45 \pm 0.74$ & $91.31 \pm 0.78$  & $89.67 \pm 0.62$  &$89.58 \pm 0.64$   & $95.81 \pm 0.12$ & $95.76 \pm 0.12$ & $68.49 \pm 3.12$ & $68.29 \pm 3.21$ \\

\textit{TDOST-LLM+Temporal}   & \cellcolor[HTML]{caebc0} $\textbf{91.61} \pm \textbf{0.68} $  & \cellcolor[HTML]{caebc0} $\textbf{91.45} \pm \textbf{0.70}$ & \cellcolor[HTML]{caebc0}  $\textbf{89.58} \pm \textbf{1.14}$ & \cellcolor[HTML]{caebc0} $\textbf{89.49} \pm \textbf{1.06}$  & \cellcolor[HTML]{caebc0} $96.14 \pm 0.06$ & \cellcolor[HTML]{caebc0} $96.10 \pm 0.05$ & $70.16 \pm 4.67$ & $69.99 \pm 4.73$  \\

\midrule
\textit{DeepCASAS}    & $89.32 \pm 0.47$  &  $88.99 \pm 0.44$  & $85.85 \pm 1.05$ & $85.25 \pm 1.38$  & $\textbf{96.50} \pm \textbf{0.24}$ &  $\textbf{96.46} \pm \textbf{0.22}$ & $73.44 \pm 2.25$ & $73.15 \pm 2.29$\\

\bottomrule
\end{tabular}
}
\caption{Performance of the TDOST versions in the standard, supervised HAR setup, i.e., without transfer between source and target homes. This setting serves as baseline comparison for the predominant training approach where models have access to labeled sample data from each individual target home. 
The overall best results in every configuration are boldfaced, whereas, the best-performing TDOST version is highlighted in the green color. 
In the Milan, Cairo, and Kyoto7 datasets, TDOST outperforms the DeepCASAS setup, whereas, in the Aruba dataset, it performs comparably. 
These results demonstrate that TDOST-based modeling can also be used for standard, non-transfer application scenarios.
}
\label{tab:har_tdost}
\end{table}

\section{Summary and Discussion}
\label{sec:discussion}
Developing a HAR system for each individual smart home is resource intensive requiring large amounts of labeled data instances to learn robust models that are specific to the given home and to its residents \cite{hiremath2022bootstrapping}. 
We presented a novel approach to layout-agnostic HAR that allows for the transfer of activity recognition systems across homes thereby not requiring adaptation nor retraining for target scenarios.
The key to our approach lies in the automated conversion of sensor event triggers into textual descriptions that capture relevant contextual information, which allows for the desired generalization -- TDOST.
Through our experimental evaluation, we demonstrated the excellent generalization performance of HAR systems using TDOST.

\subsection{Components of TDOST-based HAR}
\label{sec:discussion:limitations}

In our current approach, we use two variants of Sentence Transformers:
\textit{i)} the DistilRoBERTa \cite{liu2019roberta, Sanh2019DistilBERTAD} ; and 
\textit{ii)} Sentence-T5\cite{ni2021sentence} as encoders to learn the required embedding space that facilitates the desired generalization. 
As classifiers we use \textit{i)} Conv-BiLSTM; and \textit{ii)} Bi-LSTM, which is in line with the state-of-the-art in HAR for smart homes \cite{liciotti2020sequential}. 

The aforementioned choices are reasonable as evidenced by the results of our experimental evaluation.
Yet, a large number of alternative sentence or text encoders is being developed (e.g., \cite{wolf2019huggingface,jain2022hugging}) and more extensive explorations of their suitability may lead to further improvements in our setting. 
The same holds for the classification backend for our HAR system.
Focusing on the general proof of concept for our novel modeling and deployment approach in this paper, we plan to explore the effects of the individual components of our method in future work.

Our approach specifically focuses on transfer across smart homes with differing floorplans (layouts) and sensor specifications.
As demonstrated, the transfer performs very well -- yet, it is worth reminding that a successful transfer is based on a shared, standard set of activities.
The activities in our evaluation are similar, if not the same, across all houses.
While this could be interpreted as a limitation, we argue that the chosen--standard--set of activities covers the most relevant practical activities in a smart home and thus our results are of importance for practical applications. 
Yet, future work will need to focus on transfer at the activity level as well.
One exemplary way to accommodate for activity transfer as well is to incorporate few-shot learning procedures \cite{wang2020generalizing, parnami2022learning, sung2018learning} and metric learning approaches \cite{snell2017prototypical} to identify the activities specific to a given target home.

\subsection{Collate Knowledge from Multiple Source Homes}
\label{sec:discussion:transfer-knowledge}
We evaluated the effectiveness of TDOST-based HAR systems through experiments based on pairs of source--target smart homes.
It is reasonable to extend our approach towards training from multiple source homes and, as such, to incorporate knowledge from a larger variety of contexts.
This would allow us to establish a larger common knowledge base available through the datasets collected across various homes, and to learn a more general embedding space.
Learning from a broader base as a possible next step for our work is motivated by our experimental evaluation where 
we observed that some source homes serve as seemingly better base as they led to somewhat better recognition capabilities for a target home.
This indicates that--beyond the general proof of concept given in this paper where we did not focus on selecting specific source homes--a less arbitrary choice of source homes, or even a broadening to more than one, might be beneficial.
Future work could investigate how to choose the source home(s), for example, through higher-level comparison and analysis of factors that have a positive impact on the downstream performance.

Drawing from the broader field of 
transfer learning would guide us in identifying optimal source data that can be used to transfer knowledge \cite{rashidi2011domain}.
In line with the core of our approach, it will be interesting to understand what constitutes a relevant commonality across smart homes, using natural language as the primary modality. 
Knowledge from here could be used to automatically bootstrap an ontology of smart homes that could be used to guide the training procedure -- yet, no detailed target home knowledge must be used as this would violate our overarching motivation for layout-agnostic HAR.


\subsection{Explainability of HAR Systems}
\label{sec:discussion:xai}
Recent works have explored the use of explainability to build trust and fairness in machine learning models \cite{angelov2021explainable}. 
In the realm of smart home research as well, explainable AI (XAI) techniques have been used to understand the reasoning behind the predictions made for sensor event sequences \cite{das2023explainable}. 
Such methods specifically utilize language to explain the decisions of the classifier, thus making sense of the black-box predictions obtained from classification backends. 

Given that our approach is based on textual input, a potential extension could incorporate the aforementioned language-based explanations to 
understand what portions of our textual descriptions led to a certain prediction, which, in turn, would allow us to further optimize TDOST. 

\subsection{Maintenance of TDOST based HAR Systems}
\label{sec:discussion:maintenance}
``Life is every changing'' \cite{hiremath2023lifespan} and with that, it is unreasonable to assume smart homes and their HAR systems as static entities that, once trained are deployed and remain ``as is'' throughout the lifetime of a house.
Aiming to automate the entirety of TDOST based HAR systems for life-long deployments, our goal is to investigate ways to achieve the flexibility of generalized deployments along with adaptability throughout the entire lifespan of a smart home and its HAR system.

Starting with obtaining a home's layout that is used for the generation of our textual descriptions of sensor event triggers: our current approach relies on a manual procedure to generate the mapping (\ref{appendix_tdost_basic}). Yet, it is reasonable to assume that modern, pretrained vision-language models (e.g., \cite{liu2024visual, achiam2023gpt}) could be used to automate this process, which would further lower the barrier for adopting our approach.

Once the HAR systems are deployed, the sensor mappings and specifications could be automatically updated as things (may) change: New appliances and new sensors might be integrated; new furniture might be added and old furniture might be removed; even structural changes to the house might happen. Automatically maintaining the meta-information about a target house and, in turn, updating the core of TDOST mapping is desirable and possible -- yet, it still would not require retraining the HAR system itself, which is the key advantage of our approach.


\section{Conclusion}
Developing activity recognition systems for smart homes can be expensive and cumbersome, given the costs (both manual and computational) associated with data collection, annotation, model training, and deployment.
We developed an approach that facilitates training activity recognition systems for smart homes that are layout-agnostic, i.e., they can be easily deployed in homes that were not included for model training. 
The key for this generalization capabilities is a novel textual representation of raw sensor data--TDOST--that adds readily available context information about a target house to the sensor trigger events.
The results of our experimental evaluation on standard CASAS benchmarks demonstrate the general effectiveness of TDOST-based HAR systems.






\bibliographystyle{ACM-Reference-Format}
\bibliography{refs}


\begin{thebibliography}{100}


\ifx \showCODEN    \undefined \def \showCODEN     #1{\unskip}     \fi
\ifx \showDOI      \undefined \def \showDOI       #1{#1}\fi
\ifx \showISBNx    \undefined \def \showISBNx     #1{\unskip}     \fi
\ifx \showISBNxiii \undefined \def \showISBNxiii  #1{\unskip}     \fi
\ifx \showISSN     \undefined \def \showISSN      #1{\unskip}     \fi
\ifx \showLCCN     \undefined \def \showLCCN      #1{\unskip}     \fi
\ifx \shownote     \undefined \def \shownote      #1{#1}          \fi
\ifx \showarticletitle \undefined \def \showarticletitle #1{#1}   \fi
\ifx \showURL      \undefined \def \showURL       {\relax}        \fi
\providecommand\bibfield[2]{#2}
\providecommand\bibinfo[2]{#2}
\providecommand\natexlab[1]{#1}
\providecommand\showeprint[2][]{arXiv:#2}

\bibitem[Achiam et~al\mbox{.}(2023)]%
        {achiam2023gpt}
\bibfield{author}{\bibinfo{person}{Josh Achiam}, \bibinfo{person}{Steven Adler}, \bibinfo{person}{Sandhini Agarwal}, \bibinfo{person}{Lama Ahmad}, \bibinfo{person}{Ilge Akkaya}, \bibinfo{person}{Florencia~Leoni Aleman}, \bibinfo{person}{Diogo Almeida}, \bibinfo{person}{Janko Altenschmidt}, \bibinfo{person}{Sam Altman}, \bibinfo{person}{Shyamal Anadkat}, {et~al\mbox{.}}} \bibinfo{year}{2023}\natexlab{}.
\newblock \showarticletitle{Gpt-4 technical report}.
\newblock \bibinfo{journal}{\emph{arXiv preprint arXiv:2303.08774}} (\bibinfo{year}{2023}).
\newblock


\bibitem[Adaimi and Thomaz(2019)]%
        {adaimi2019leveraging}
\bibfield{author}{\bibinfo{person}{Rebecca Adaimi} {and} \bibinfo{person}{Edison Thomaz}.} \bibinfo{year}{2019}\natexlab{}.
\newblock \showarticletitle{Leveraging active learning and conditional mutual information to minimize data annotation in human activity recognition}.
\newblock \bibinfo{journal}{\emph{Proceedings of the ACM on Interactive, Mobile, Wearable and Ubiquitous Technologies}} \bibinfo{volume}{3}, \bibinfo{number}{3} (\bibinfo{year}{2019}), \bibinfo{pages}{1--23}.
\newblock


\bibitem[Agarwal et~al\mbox{.}(2021)]%
        {agarwal2021transfer}
\bibfield{author}{\bibinfo{person}{Nidhi Agarwal}, \bibinfo{person}{Akanksha Sondhi}, \bibinfo{person}{Khyati Chopra}, {and} \bibinfo{person}{Ghanapriya Singh}.} \bibinfo{year}{2021}\natexlab{}.
\newblock \showarticletitle{Transfer learning: Survey and classification}.
\newblock \bibinfo{journal}{\emph{Smart Innovations in Communication and Computational Sciences: Proceedings of ICSICCS 2020}} (\bibinfo{year}{2021}), \bibinfo{pages}{145--155}.
\newblock


\bibitem[Al~Machot et~al\mbox{.}(2020)]%
        {al2020zero}
\bibfield{author}{\bibinfo{person}{Fadi Al~Machot}, \bibinfo{person}{Mohammed R.~Elkobaisi}, {and} \bibinfo{person}{Kyandoghere Kyamakya}.} \bibinfo{year}{2020}\natexlab{}.
\newblock \showarticletitle{Zero-shot human activity recognition using non-visual sensors}.
\newblock \bibinfo{journal}{\emph{Sensors}} \bibinfo{volume}{20}, \bibinfo{number}{3} (\bibinfo{year}{2020}), \bibinfo{pages}{825}.
\newblock


\bibitem[Alam et~al\mbox{.}(2012)]%
        {alam2012review}
\bibfield{author}{\bibinfo{person}{Muhammad~Raisul Alam}, \bibinfo{person}{Mamun Bin~Ibne Reaz}, {and} \bibinfo{person}{Mohd Alauddin~Mohd Ali}.} \bibinfo{year}{2012}\natexlab{}.
\newblock \showarticletitle{A review of smart homes—Past, present, and future}.
\newblock \bibinfo{journal}{\emph{IEEE transactions on systems, man, and cybernetics, part C (applications and reviews)}} \bibinfo{volume}{42}, \bibinfo{number}{6} (\bibinfo{year}{2012}), \bibinfo{pages}{1190--1203}.
\newblock


\bibitem[Alemdar et~al\mbox{.}(2013)]%
        {alemdar2013aras}
\bibfield{author}{\bibinfo{person}{Hande Alemdar}, \bibinfo{person}{Halil Ertan}, \bibinfo{person}{Ozlem~Durmaz Incel}, {and} \bibinfo{person}{Cem Ersoy}.} \bibinfo{year}{2013}\natexlab{}.
\newblock \showarticletitle{ARAS human activity datasets in multiple homes with multiple residents}. In \bibinfo{booktitle}{\emph{2013 7th International Conference on Pervasive Computing Technologies for Healthcare and Workshops}}. IEEE, \bibinfo{pages}{232--235}.
\newblock


\bibitem[Allik et~al\mbox{.}(2019)]%
        {allik2019optimization}
\bibfield{author}{\bibinfo{person}{Ardo Allik}, \bibinfo{person}{Kristjan Pilt}, \bibinfo{person}{Deniss Karai}, \bibinfo{person}{Ivo Fridolin}, \bibinfo{person}{Mairo Leier}, {and} \bibinfo{person}{Gert Jervan}.} \bibinfo{year}{2019}\natexlab{}.
\newblock \showarticletitle{Optimization of physical activity recognition for real-time wearable systems: effect of window length, sampling frequency and number of features}.
\newblock \bibinfo{journal}{\emph{Applied Sciences}} \bibinfo{volume}{9}, \bibinfo{number}{22} (\bibinfo{year}{2019}), \bibinfo{pages}{4833}.
\newblock


\bibitem[Aminikhanghahi and Cook(2017)]%
        {aminikhanghahi2017using}
\bibfield{author}{\bibinfo{person}{Samaneh Aminikhanghahi} {and} \bibinfo{person}{Diane~J Cook}.} \bibinfo{year}{2017}\natexlab{}.
\newblock \showarticletitle{Using change point detection to automate daily activity segmentation}. In \bibinfo{booktitle}{\emph{2017 IEEE International Conference on Pervasive Computing and Communications Workshops (PerCom Workshops)}}. IEEE, \bibinfo{pages}{262--267}.
\newblock


\bibitem[Aminikhanghahi et~al\mbox{.}(2018)]%
        {aminikhanghahi2018real}
\bibfield{author}{\bibinfo{person}{Samaneh Aminikhanghahi}, \bibinfo{person}{Tinghui Wang}, {and} \bibinfo{person}{Diane~J Cook}.} \bibinfo{year}{2018}\natexlab{}.
\newblock \showarticletitle{Real-time change point detection with application to smart home time series data}.
\newblock \bibinfo{journal}{\emph{IEEE Transactions on Knowledge and Data Engineering}} \bibinfo{volume}{31}, \bibinfo{number}{5} (\bibinfo{year}{2018}), \bibinfo{pages}{1010--1023}.
\newblock


\bibitem[Angelov et~al\mbox{.}(2021)]%
        {angelov2021explainable}
\bibfield{author}{\bibinfo{person}{Plamen~P Angelov}, \bibinfo{person}{Eduardo~A Soares}, \bibinfo{person}{Richard Jiang}, \bibinfo{person}{Nicholas~I Arnold}, {and} \bibinfo{person}{Peter~M Atkinson}.} \bibinfo{year}{2021}\natexlab{}.
\newblock \showarticletitle{Explainable artificial intelligence: an analytical review}.
\newblock \bibinfo{journal}{\emph{Wiley Interdisciplinary Reviews: Data Mining and Knowledge Discovery}} \bibinfo{volume}{11}, \bibinfo{number}{5} (\bibinfo{year}{2021}), \bibinfo{pages}{e1424}.
\newblock


\bibitem[Bascol et~al\mbox{.}(2019)]%
        {bascol2019improving}
\bibfield{author}{\bibinfo{person}{Kevin Bascol}, \bibinfo{person}{R{\'e}mi Emonet}, {and} \bibinfo{person}{Elisa Fromont}.} \bibinfo{year}{2019}\natexlab{}.
\newblock \showarticletitle{Improving domain adaptation by source selection}. In \bibinfo{booktitle}{\emph{2019 IEEE International Conference on Image Processing (ICIP)}}. IEEE, \bibinfo{pages}{3043--3047}.
\newblock


\bibitem[Bettadapura et~al\mbox{.}(2013)]%
        {bettadapura2013augmenting}
\bibfield{author}{\bibinfo{person}{Vinay Bettadapura}, \bibinfo{person}{Grant Schindler}, \bibinfo{person}{Thomas Pl{\"o}tz}, {and} \bibinfo{person}{Irfan Essa}.} \bibinfo{year}{2013}\natexlab{}.
\newblock \showarticletitle{Augmenting bag-of-words: Data-driven discovery of temporal and structural information for activity recognition}. In \bibinfo{booktitle}{\emph{Proceedings of the IEEE Conference on Computer Vision and Pattern Recognition}}. \bibinfo{pages}{2619--2626}.
\newblock


\bibitem[Boovaraghavan et~al\mbox{.}(2023)]%
        {boovaraghavan2023tao}
\bibfield{author}{\bibinfo{person}{Sudershan Boovaraghavan}, \bibinfo{person}{Prasoon Patidar}, {and} \bibinfo{person}{Yuvraj Agarwal}.} \bibinfo{year}{2023}\natexlab{}.
\newblock \showarticletitle{TAO: Context Detection from Daily Activity Patterns Using Temporal Analysis and Ontology}.
\newblock \bibinfo{journal}{\emph{Proceedings of the ACM on Interactive, Mobile, Wearable and Ubiquitous Technologies}} \bibinfo{volume}{7}, \bibinfo{number}{3} (\bibinfo{year}{2023}), \bibinfo{pages}{1--32}.
\newblock


\bibitem[Bouchabou et~al\mbox{.}(2021a)]%
        {bouchabou2021human}
\bibfield{author}{\bibinfo{person}{Damien Bouchabou}, \bibinfo{person}{Christophe Lohr}, \bibinfo{person}{Ioannis Kanellos}, {and} \bibinfo{person}{Sao~Mai Nguyen}.} \bibinfo{year}{2021}\natexlab{a}.
\newblock \showarticletitle{Human Activity Recognition (HAR) in smart homes}.
\newblock \bibinfo{journal}{\emph{arXiv preprint arXiv:2112.11232}} (\bibinfo{year}{2021}).
\newblock


\bibitem[Bouchabou et~al\mbox{.}(2021b)]%
        {bouchabou2021fully}
\bibfield{author}{\bibinfo{person}{Damien Bouchabou}, \bibinfo{person}{Sao~Mai Nguyen}, \bibinfo{person}{Christophe Lohr}, \bibinfo{person}{Benoit Leduc}, {and} \bibinfo{person}{Ioannis Kanellos}.} \bibinfo{year}{2021}\natexlab{b}.
\newblock \showarticletitle{Fully convolutional network bootstrapped by word encoding and embedding for activity recognition in smart homes}. In \bibinfo{booktitle}{\emph{Deep Learning for Human Activity Recognition: Second International Workshop, DL-HAR 2020, Held in Conjunction with IJCAI-PRICAI 2020, Kyoto, Japan, January 8, 2021, Proceedings 2}}. Springer, \bibinfo{pages}{111--125}.
\newblock


\bibitem[Bouchabou et~al\mbox{.}(2021c)]%
        {bouchabou2021survey}
\bibfield{author}{\bibinfo{person}{Damien Bouchabou}, \bibinfo{person}{Sao~Mai Nguyen}, \bibinfo{person}{Christophe Lohr}, \bibinfo{person}{Benoit LeDuc}, {and} \bibinfo{person}{Ioannis Kanellos}.} \bibinfo{year}{2021}\natexlab{c}.
\newblock \showarticletitle{A survey of human activity recognition in smart homes based on IoT sensors algorithms: Taxonomies, challenges, and opportunities with deep learning}.
\newblock \bibinfo{journal}{\emph{Sensors}} \bibinfo{volume}{21}, \bibinfo{number}{18} (\bibinfo{year}{2021}), \bibinfo{pages}{6037}.
\newblock


\bibitem[Bouchabou et~al\mbox{.}(2021d)]%
        {bouchabou2021using}
\bibfield{author}{\bibinfo{person}{Damien Bouchabou}, \bibinfo{person}{Sao~Mai Nguyen}, \bibinfo{person}{Christophe Lohr}, \bibinfo{person}{Benoit LeDuc}, {and} \bibinfo{person}{Ioannis Kanellos}.} \bibinfo{year}{2021}\natexlab{d}.
\newblock \showarticletitle{Using language model to bootstrap human activity recognition ambient sensors based in smart homes}.
\newblock \bibinfo{journal}{\emph{Electronics}} \bibinfo{volume}{10}, \bibinfo{number}{20} (\bibinfo{year}{2021}), \bibinfo{pages}{2498}.
\newblock


\bibitem[Chan et~al\mbox{.}(2008)]%
        {chan2008review}
\bibfield{author}{\bibinfo{person}{Marie Chan}, \bibinfo{person}{Daniel Est{\`e}ve}, \bibinfo{person}{Christophe Escriba}, {and} \bibinfo{person}{Eric Campo}.} \bibinfo{year}{2008}\natexlab{}.
\newblock \showarticletitle{A review of smart homes—Present state and future challenges}.
\newblock \bibinfo{journal}{\emph{Computer methods and programs in biomedicine}} \bibinfo{volume}{91}, \bibinfo{number}{1} (\bibinfo{year}{2008}), \bibinfo{pages}{55--81}.
\newblock


\bibitem[Chang et~al\mbox{.}(2017)]%
        {chang2017unsupervised}
\bibfield{author}{\bibinfo{person}{Hang Chang}, \bibinfo{person}{Ju Han}, \bibinfo{person}{Cheng Zhong}, \bibinfo{person}{Antoine~M Snijders}, {and} \bibinfo{person}{Jian-Hua Mao}.} \bibinfo{year}{2017}\natexlab{}.
\newblock \showarticletitle{Unsupervised transfer learning via multi-scale convolutional sparse coding for biomedical applications}.
\newblock \bibinfo{journal}{\emph{IEEE transactions on pattern analysis and machine intelligence}} \bibinfo{volume}{40}, \bibinfo{number}{5} (\bibinfo{year}{2017}), \bibinfo{pages}{1182--1194}.
\newblock


\bibitem[Chang et~al\mbox{.}(2020)]%
        {chang2020systematic}
\bibfield{author}{\bibinfo{person}{Youngjae Chang}, \bibinfo{person}{Akhil Mathur}, \bibinfo{person}{Anton Isopoussu}, \bibinfo{person}{Junehwa Song}, {and} \bibinfo{person}{Fahim Kawsar}.} \bibinfo{year}{2020}\natexlab{}.
\newblock \showarticletitle{A systematic study of unsupervised domain adaptation for robust human-activity recognition}.
\newblock \bibinfo{journal}{\emph{Proceedings of the ACM on Interactive, Mobile, Wearable and Ubiquitous Technologies}} \bibinfo{volume}{4}, \bibinfo{number}{1} (\bibinfo{year}{2020}), \bibinfo{pages}{1--30}.
\newblock


\bibitem[Chatting(2023)]%
        {chatting2023automated}
\bibfield{author}{\bibinfo{person}{David Chatting}.} \bibinfo{year}{2023}\natexlab{}.
\newblock \showarticletitle{Automated Indifference}.
\newblock \bibinfo{journal}{\emph{Interactions}} \bibinfo{volume}{30}, \bibinfo{number}{2} (\bibinfo{year}{2023}), \bibinfo{pages}{22--26}.
\newblock


\bibitem[Chen et~al\mbox{.}(2023)]%
        {chen2023leveraging}
\bibfield{author}{\bibinfo{person}{Hui Chen}, \bibinfo{person}{Charles Gouin-Vallerand}, \bibinfo{person}{K{\'e}vin Bouchard}, \bibinfo{person}{S{\'e}bastien Gaboury}, \bibinfo{person}{M{\'e}lanie Couture}, \bibinfo{person}{Nathalie Bier}, {and} \bibinfo{person}{Sylvain Giroux}.} \bibinfo{year}{2023}\natexlab{}.
\newblock \showarticletitle{Leveraging Self-Supervised Learning for Human Activity Recognition with Ambient Sensors}. In \bibinfo{booktitle}{\emph{Proceedings of the 2023 ACM Conference on Information Technology for Social Good}}. \bibinfo{pages}{324--332}.
\newblock


\bibitem[Chen et~al\mbox{.}(2024)]%
        {chen2024enhancing}
\bibfield{author}{\bibinfo{person}{Hui Chen}, \bibinfo{person}{Charles Gouin-Vallerand}, \bibinfo{person}{K{\'e}vin Bouchard}, \bibinfo{person}{S{\'e}bastien Gaboury}, \bibinfo{person}{M{\'e}lanie Couture}, \bibinfo{person}{Nathalie Bier}, {and} \bibinfo{person}{Sylvain Giroux}.} \bibinfo{year}{2024}\natexlab{}.
\newblock \showarticletitle{Enhancing Human Activity Recognition in Smart Homes with Self-Supervised Learning and Self-Attention}.
\newblock \bibinfo{journal}{\emph{Sensors}} \bibinfo{volume}{24}, \bibinfo{number}{3} (\bibinfo{year}{2024}), \bibinfo{pages}{884}.
\newblock


\bibitem[Chen and Nugent(2009)]%
        {chen2009ontology}
\bibfield{author}{\bibinfo{person}{Liming Chen} {and} \bibinfo{person}{Chris Nugent}.} \bibinfo{year}{2009}\natexlab{}.
\newblock \showarticletitle{Ontology-based activity recognition in intelligent pervasive environments}.
\newblock \bibinfo{journal}{\emph{International Journal of Web Information Systems}} \bibinfo{volume}{5}, \bibinfo{number}{4} (\bibinfo{year}{2009}), \bibinfo{pages}{410--430}.
\newblock


\bibitem[Chen and Nugent(2019)]%
        {chen2019human}
\bibfield{author}{\bibinfo{person}{Liming Chen} {and} \bibinfo{person}{Chris~D Nugent}.} \bibinfo{year}{2019}\natexlab{}.
\newblock \bibinfo{booktitle}{\emph{Human activity recognition and behaviour analysis}}.
\newblock \bibinfo{publisher}{Springer}.
\newblock


\bibitem[Chiang and Hsu(2012)]%
        {chiang2012knowledge}
\bibfield{author}{\bibinfo{person}{Yi-ting Chiang} {and} \bibinfo{person}{Jane Yung-jen Hsu}.} \bibinfo{year}{2012}\natexlab{}.
\newblock \showarticletitle{Knowledge transfer in activity recognition using sensor profile}. In \bibinfo{booktitle}{\emph{2012 9th International Conference on Ubiquitous Intelligence and Computing and 9th International Conference on Autonomic and Trusted Computing}}. IEEE, \bibinfo{pages}{180--187}.
\newblock


\bibitem[Cook(2010)]%
        {cook2010learning}
\bibfield{author}{\bibinfo{person}{Diane~J Cook}.} \bibinfo{year}{2010}\natexlab{}.
\newblock \showarticletitle{Learning setting-generalized activity models for smart spaces}.
\newblock \bibinfo{journal}{\emph{IEEE intelligent systems}} \bibinfo{volume}{2010}, \bibinfo{number}{99} (\bibinfo{year}{2010}), \bibinfo{pages}{1}.
\newblock


\bibitem[Cook(2020)]%
        {cook2020activity}
\bibfield{author}{\bibinfo{person}{Diane~J Cook}.} \bibinfo{year}{2020}\natexlab{}.
\newblock \showarticletitle{AL Activity Learning—Smart Home}.
\newblock \bibinfo{journal}{\emph{Retrieved April}}  \bibinfo{volume}{28} (\bibinfo{year}{2020}), \bibinfo{pages}{2023}.
\newblock


\bibitem[Cook et~al\mbox{.}(2012)]%
        {cook2012casas}
\bibfield{author}{\bibinfo{person}{Diane~J Cook}, \bibinfo{person}{Aaron~S Crandall}, \bibinfo{person}{Brian~L Thomas}, {and} \bibinfo{person}{Narayanan~C Krishnan}.} \bibinfo{year}{2012}\natexlab{}.
\newblock \showarticletitle{CASAS: A smart home in a box}.
\newblock \bibinfo{journal}{\emph{Computer}} \bibinfo{volume}{46}, \bibinfo{number}{7} (\bibinfo{year}{2012}), \bibinfo{pages}{62--69}.
\newblock


\bibitem[Cook et~al\mbox{.}(2013)]%
        {cook2013activity}
\bibfield{author}{\bibinfo{person}{Diane~J Cook}, \bibinfo{person}{Narayanan~C Krishnan}, {and} \bibinfo{person}{Parisa Rashidi}.} \bibinfo{year}{2013}\natexlab{}.
\newblock \showarticletitle{Activity discovery and activity recognition: A new partnership}.
\newblock \bibinfo{journal}{\emph{IEEE transactions on cybernetics}} \bibinfo{volume}{43}, \bibinfo{number}{3} (\bibinfo{year}{2013}), \bibinfo{pages}{820--828}.
\newblock


\bibitem[Cook et~al\mbox{.}(2003)]%
        {cook2003mavhome}
\bibfield{author}{\bibinfo{person}{Diane~J Cook}, \bibinfo{person}{Michael Youngblood}, \bibinfo{person}{Edwin~O Heierman}, \bibinfo{person}{Karthik Gopalratnam}, \bibinfo{person}{Sira Rao}, \bibinfo{person}{Andrey Litvin}, {and} \bibinfo{person}{Farhan Khawaja}.} \bibinfo{year}{2003}\natexlab{}.
\newblock \showarticletitle{MavHome: An agent-based smart home}. In \bibinfo{booktitle}{\emph{Proceedings of the First IEEE International Conference on Pervasive Computing and Communications, 2003.(PerCom 2003).}} IEEE, \bibinfo{pages}{521--524}.
\newblock


\bibitem[Csurka(2017)]%
        {csurka2017domain}
\bibfield{author}{\bibinfo{person}{Gabriela Csurka}.} \bibinfo{year}{2017}\natexlab{}.
\newblock \showarticletitle{Domain adaptation for visual applications: A comprehensive survey}.
\newblock \bibinfo{journal}{\emph{arXiv preprint arXiv:1702.05374}} (\bibinfo{year}{2017}).
\newblock


\bibitem[Dahmen et~al\mbox{.}(2017)]%
        {dahmen2017activity}
\bibfield{author}{\bibinfo{person}{Jessamyn Dahmen}, \bibinfo{person}{Brian~L Thomas}, \bibinfo{person}{Diane~J Cook}, {and} \bibinfo{person}{Xiaobo Wang}.} \bibinfo{year}{2017}\natexlab{}.
\newblock \showarticletitle{Activity learning as a foundation for security monitoring in smart homes}.
\newblock \bibinfo{journal}{\emph{Sensors}} \bibinfo{volume}{17}, \bibinfo{number}{4} (\bibinfo{year}{2017}), \bibinfo{pages}{737}.
\newblock


\bibitem[Dang et~al\mbox{.}(2020)]%
        {dang2020sensor}
\bibfield{author}{\bibinfo{person}{L~Minh Dang}, \bibinfo{person}{Kyungbok Min}, \bibinfo{person}{Hanxiang Wang}, \bibinfo{person}{Md~Jalil Piran}, \bibinfo{person}{Cheol~Hee Lee}, {and} \bibinfo{person}{Hyeonjoon Moon}.} \bibinfo{year}{2020}\natexlab{}.
\newblock \showarticletitle{Sensor-based and vision-based human activity recognition: A comprehensive survey}.
\newblock \bibinfo{journal}{\emph{Pattern Recognition}}  \bibinfo{volume}{108} (\bibinfo{year}{2020}), \bibinfo{pages}{107561}.
\newblock


\bibitem[Das et~al\mbox{.}(2023)]%
        {das2023explainable}
\bibfield{author}{\bibinfo{person}{Devleena Das}, \bibinfo{person}{Yasutaka Nishimura}, \bibinfo{person}{Rajan~P Vivek}, \bibinfo{person}{Naoto Takeda}, \bibinfo{person}{Sean~T Fish}, \bibinfo{person}{Thomas Ploetz}, {and} \bibinfo{person}{Sonia Chernova}.} \bibinfo{year}{2023}\natexlab{}.
\newblock \showarticletitle{Explainable activity recognition for smart home systems}.
\newblock \bibinfo{journal}{\emph{ACM Transactions on Interactive Intelligent Systems}} \bibinfo{volume}{13}, \bibinfo{number}{2} (\bibinfo{year}{2023}), \bibinfo{pages}{1--39}.
\newblock


\bibitem[Day and Khoshgoftaar(2017)]%
        {day2017survey}
\bibfield{author}{\bibinfo{person}{Oscar Day} {and} \bibinfo{person}{Taghi~M Khoshgoftaar}.} \bibinfo{year}{2017}\natexlab{}.
\newblock \showarticletitle{A survey on heterogeneous transfer learning}.
\newblock \bibinfo{journal}{\emph{Journal of Big Data}}  \bibinfo{volume}{4} (\bibinfo{year}{2017}), \bibinfo{pages}{1--42}.
\newblock


\bibitem[Dhekane and Ploetz(2024)]%
        {dhekane2024transfer}
\bibfield{author}{\bibinfo{person}{Sourish~Gunesh Dhekane} {and} \bibinfo{person}{Thomas Ploetz}.} \bibinfo{year}{2024}\natexlab{}.
\newblock \showarticletitle{Transfer Learning in Human Activity Recognition: A Survey}.
\newblock \bibinfo{journal}{\emph{arXiv preprint arXiv:2401.10185}} (\bibinfo{year}{2024}).
\newblock


\bibitem[Fahad et~al\mbox{.}(2014)]%
        {fahad2014activity}
\bibfield{author}{\bibinfo{person}{Labiba~Gillani Fahad}, \bibinfo{person}{Syed~Fahad Tahir}, {and} \bibinfo{person}{Muttukrishnan Rajarajan}.} \bibinfo{year}{2014}\natexlab{}.
\newblock \showarticletitle{Activity recognition in smart homes using clustering based classification}. In \bibinfo{booktitle}{\emph{2014 22nd International conference on pattern recognition}}. IEEE, \bibinfo{pages}{1348--1353}.
\newblock


\bibitem[Farahani et~al\mbox{.}(2021)]%
        {farahani2021brief}
\bibfield{author}{\bibinfo{person}{Abolfazl Farahani}, \bibinfo{person}{Sahar Voghoei}, \bibinfo{person}{Khaled Rasheed}, {and} \bibinfo{person}{Hamid~R Arabnia}.} \bibinfo{year}{2021}\natexlab{}.
\newblock \showarticletitle{A brief review of domain adaptation}.
\newblock \bibinfo{journal}{\emph{Advances in data science and information engineering: proceedings from ICDATA 2020 and IKE 2020}} (\bibinfo{year}{2021}), \bibinfo{pages}{877--894}.
\newblock


\bibitem[Fida et~al\mbox{.}(2014)]%
        {fida2014effect}
\bibfield{author}{\bibinfo{person}{Benish Fida}, \bibinfo{person}{Ivan Bernabucci}, \bibinfo{person}{Daniele Bibbo}, \bibinfo{person}{Silvia Conforto}, \bibinfo{person}{Antonino Proto}, {and} \bibinfo{person}{Maurizio Schmid}.} \bibinfo{year}{2014}\natexlab{}.
\newblock \showarticletitle{The effect of window length on the classification of dynamic activities through a single accelerometer}. In \bibinfo{booktitle}{\emph{Biomedical Engineering. In Proceedings of the IASTED International Conference Biomedical Engineering (BioMed), Zurich, Switerzland}}. \bibinfo{pages}{23--25}.
\newblock


\bibitem[Fida et~al\mbox{.}(2015)]%
        {fida2015varying}
\bibfield{author}{\bibinfo{person}{Benish Fida}, \bibinfo{person}{Ivan Bernabucci}, \bibinfo{person}{Daniele Bibbo}, \bibinfo{person}{Silvia Conforto}, {and} \bibinfo{person}{Maurizio Schmid}.} \bibinfo{year}{2015}\natexlab{}.
\newblock \showarticletitle{Varying behavior of different window sizes on the classification of static and dynamic physical activities from a single accelerometer}.
\newblock \bibinfo{journal}{\emph{Medical engineering \& physics}} \bibinfo{volume}{37}, \bibinfo{number}{7} (\bibinfo{year}{2015}), \bibinfo{pages}{705--711}.
\newblock


\bibitem[Ghods and Cook(2019)]%
        {ghods2019activity2vec}
\bibfield{author}{\bibinfo{person}{Alireza Ghods} {and} \bibinfo{person}{Diane~J Cook}.} \bibinfo{year}{2019}\natexlab{}.
\newblock \showarticletitle{Activity2vec: Learning adl embeddings from sensor data with a sequence-to-sequence model}.
\newblock \bibinfo{journal}{\emph{arXiv preprint arXiv:1907.05597}} (\bibinfo{year}{2019}).
\newblock


\bibitem[Gokaslan and Cohen(2019)]%
        {Gokaslan2019OpenWeb}
\bibfield{author}{\bibinfo{person}{Aaron Gokaslan} {and} \bibinfo{person}{Vanya Cohen}.} \bibinfo{year}{2019}\natexlab{}.
\newblock \bibinfo{title}{OpenWebText Corpus}.
\newblock \bibinfo{howpublished}{\url{http://Skylion007.github.io/OpenWebTextCorpus}}.
\newblock


\bibitem[Guo et~al\mbox{.}(2018)]%
        {guo2018deep}
\bibfield{author}{\bibinfo{person}{Liang Guo}, \bibinfo{person}{Yaguo Lei}, \bibinfo{person}{Saibo Xing}, \bibinfo{person}{Tao Yan}, {and} \bibinfo{person}{Naipeng Li}.} \bibinfo{year}{2018}\natexlab{}.
\newblock \showarticletitle{Deep convolutional transfer learning network: A new method for intelligent fault diagnosis of machines with unlabeled data}.
\newblock \bibinfo{journal}{\emph{IEEE Transactions on Industrial Electronics}} \bibinfo{volume}{66}, \bibinfo{number}{9} (\bibinfo{year}{2018}), \bibinfo{pages}{7316--7325}.
\newblock


\bibitem[He et~al\mbox{.}(2021)]%
        {he2021towards}
\bibfield{author}{\bibinfo{person}{Junxian He}, \bibinfo{person}{Chunting Zhou}, \bibinfo{person}{Xuezhe Ma}, \bibinfo{person}{Taylor Berg-Kirkpatrick}, {and} \bibinfo{person}{Graham Neubig}.} \bibinfo{year}{2021}\natexlab{}.
\newblock \showarticletitle{Towards a unified view of parameter-efficient transfer learning}.
\newblock \bibinfo{journal}{\emph{arXiv preprint arXiv:2110.04366}} (\bibinfo{year}{2021}).
\newblock


\bibitem[Helal et~al\mbox{.}(2005)]%
        {helal2005gator}
\bibfield{author}{\bibinfo{person}{Sumi Helal}, \bibinfo{person}{William Mann}, \bibinfo{person}{Hicham El-Zabadani}, \bibinfo{person}{Jeffrey King}, \bibinfo{person}{Youssef Kaddoura}, {and} \bibinfo{person}{Erwin Jansen}.} \bibinfo{year}{2005}\natexlab{}.
\newblock \showarticletitle{The gator tech smart house: A programmable pervasive space}.
\newblock \bibinfo{journal}{\emph{Computer}} \bibinfo{volume}{38}, \bibinfo{number}{3} (\bibinfo{year}{2005}), \bibinfo{pages}{50--60}.
\newblock


\bibitem[Hiremath et~al\mbox{.}(2022)]%
        {hiremath2022bootstrapping}
\bibfield{author}{\bibinfo{person}{Shruthi~K Hiremath}, \bibinfo{person}{Yasutaka Nishimura}, \bibinfo{person}{Sonia Chernova}, {and} \bibinfo{person}{Thomas Pl{\"o}tz}.} \bibinfo{year}{2022}\natexlab{}.
\newblock \showarticletitle{Bootstrapping human activity recognition systems for smart homes from scratch}.
\newblock \bibinfo{journal}{\emph{Proceedings of the ACM on Interactive, Mobile, Wearable and Ubiquitous Technologies}} \bibinfo{volume}{6}, \bibinfo{number}{3} (\bibinfo{year}{2022}), \bibinfo{pages}{1--27}.
\newblock


\bibitem[Hiremath and Pl{\"o}tz(2020)]%
        {hiremath2020deriving}
\bibfield{author}{\bibinfo{person}{Shruthi~K Hiremath} {and} \bibinfo{person}{Thomas Pl{\"o}tz}.} \bibinfo{year}{2020}\natexlab{}.
\newblock \showarticletitle{Deriving effective human activity recognition systems through objective task complexity assessment}.
\newblock \bibinfo{journal}{\emph{Proceedings of the ACM on Interactive, Mobile, Wearable and Ubiquitous Technologies}} \bibinfo{volume}{4}, \bibinfo{number}{4} (\bibinfo{year}{2020}), \bibinfo{pages}{1--24}.
\newblock


\bibitem[Hiremath and Pl{\"o}tz(2023)]%
        {hiremath2023lifespan}
\bibfield{author}{\bibinfo{person}{Shruthi~K Hiremath} {and} \bibinfo{person}{Thomas Pl{\"o}tz}.} \bibinfo{year}{2023}\natexlab{}.
\newblock \showarticletitle{The Lifespan of Human Activity Recognition Systems for Smart Homes}.
\newblock \bibinfo{journal}{\emph{Sensors}} \bibinfo{volume}{23}, \bibinfo{number}{18} (\bibinfo{year}{2023}), \bibinfo{pages}{7729}.
\newblock


\bibitem[Hooper et~al\mbox{.}(2012)]%
        {hooper2012french}
\bibfield{author}{\bibinfo{person}{Clare~J Hooper}, \bibinfo{person}{Anne Preston}, \bibinfo{person}{Madeline Balaam}, \bibinfo{person}{Paul Seedhouse}, \bibinfo{person}{Daniel Jackson}, \bibinfo{person}{Cuong Pham}, \bibinfo{person}{Cassim Ladha}, \bibinfo{person}{Karim Ladha}, \bibinfo{person}{Thomas Pl{\"o}tz}, {and} \bibinfo{person}{Patrick Olivier}.} \bibinfo{year}{2012}\natexlab{}.
\newblock \showarticletitle{The french kitchen: Task-based learning in an instrumented kitchen}. In \bibinfo{booktitle}{\emph{Proceedings of the 2012 ACM Conference on Ubiquitous Computing}}. \bibinfo{pages}{193--202}.
\newblock


\bibitem[Hoque and Stankovic(2012)]%
        {hoque2012aalo}
\bibfield{author}{\bibinfo{person}{Enamul Hoque} {and} \bibinfo{person}{John Stankovic}.} \bibinfo{year}{2012}\natexlab{}.
\newblock \showarticletitle{AALO: Activity recognition in smart homes using Active Learning in the presence of Overlapped activities}. In \bibinfo{booktitle}{\emph{2012 6th International Conference on Pervasive Computing Technologies for Healthcare (PervasiveHealth) and Workshops}}. IEEE, \bibinfo{pages}{139--146}.
\newblock


\bibitem[Hosna et~al\mbox{.}(2022)]%
        {hosna2022transfer}
\bibfield{author}{\bibinfo{person}{Asmaul Hosna}, \bibinfo{person}{Ethel Merry}, \bibinfo{person}{Jigmey Gyalmo}, \bibinfo{person}{Zulfikar Alom}, \bibinfo{person}{Zeyar Aung}, {and} \bibinfo{person}{Mohammad~Abdul Azim}.} \bibinfo{year}{2022}\natexlab{}.
\newblock \showarticletitle{Transfer learning: a friendly introduction}.
\newblock \bibinfo{journal}{\emph{Journal of Big Data}} \bibinfo{volume}{9}, \bibinfo{number}{1} (\bibinfo{year}{2022}), \bibinfo{pages}{102}.
\newblock


\bibitem[Hu and Yang(2011)]%
        {hu2011transfer}
\bibfield{author}{\bibinfo{person}{Derekhao Hu} {and} \bibinfo{person}{Qiang Yang}.} \bibinfo{year}{2011}\natexlab{}.
\newblock \showarticletitle{Transfer learning for activity recognition via sensor mapping}. In \bibinfo{booktitle}{\emph{Proceedings of the Twenty-Second International Joint Conference on Artificial Intelligence, Barcelona, Catalonia, Spain}}. \bibinfo{pages}{1962}.
\newblock


\bibitem[Intille et~al\mbox{.}(2004)]%
        {intille2004acquiring}
\bibfield{author}{\bibinfo{person}{Stephen~S Intille}, \bibinfo{person}{Ling Bao}, \bibinfo{person}{Emmanuel~Munguia Tapia}, {and} \bibinfo{person}{John Rondoni}.} \bibinfo{year}{2004}\natexlab{}.
\newblock \showarticletitle{Acquiring in situ training data for context-aware ubiquitous computing applications}. In \bibinfo{booktitle}{\emph{Proceedings of the SIGCHI conference on Human factors in computing systems}}. \bibinfo{pages}{1--8}.
\newblock


\bibitem[Intille et~al\mbox{.}(2005)]%
        {intille2005placelab}
\bibfield{author}{\bibinfo{person}{Stephen~S Intille}, \bibinfo{person}{Kent Larson}, \bibinfo{person}{Jennifer Beaudin}, \bibinfo{person}{E~Munguia Tapia}, \bibinfo{person}{Pallavi Kaushik}, \bibinfo{person}{Jason Nawyn}, {and} \bibinfo{person}{Thomas~J McLeish}.} \bibinfo{year}{2005}\natexlab{}.
\newblock \showarticletitle{The PlaceLab: A live-in laboratory for pervasive computing research (video)}.
\newblock \bibinfo{journal}{\emph{Proceedings of PERVASIVE 2005 Video Program}} (\bibinfo{year}{2005}).
\newblock


\bibitem[Jain(2022)]%
        {jain2022hugging}
\bibfield{author}{\bibinfo{person}{Shashank~Mohan Jain}.} \bibinfo{year}{2022}\natexlab{}.
\newblock \showarticletitle{Hugging face}.
\newblock In \bibinfo{booktitle}{\emph{Introduction to transformers for NLP: With the hugging face library and models to solve problems}}. \bibinfo{publisher}{Springer}, \bibinfo{pages}{51--67}.
\newblock


\bibitem[Khan and Roy(2018)]%
        {khan2018untran}
\bibfield{author}{\bibinfo{person}{Md~Abdullah Al~Hafiz Khan} {and} \bibinfo{person}{Nirmalya Roy}.} \bibinfo{year}{2018}\natexlab{}.
\newblock \showarticletitle{Untran: Recognizing unseen activities with unlabeled data using transfer learning}. In \bibinfo{booktitle}{\emph{2018 IEEE/ACM Third International Conference on Internet-of-Things Design and Implementation (IoTDI)}}. IEEE, \bibinfo{pages}{37--47}.
\newblock


\bibitem[Kientz et~al\mbox{.}(2008)]%
        {kientz2008georgia}
\bibfield{author}{\bibinfo{person}{Julie~A Kientz}, \bibinfo{person}{Shwetak~N Patel}, \bibinfo{person}{Brian Jones}, \bibinfo{person}{ED Price}, \bibinfo{person}{Elizabeth~D Mynatt}, {and} \bibinfo{person}{Gregory~D Abowd}.} \bibinfo{year}{2008}\natexlab{}.
\newblock \showarticletitle{The georgia tech aware home}.
\newblock In \bibinfo{booktitle}{\emph{CHI'08 extended abstracts on Human factors in computing systems}}. \bibinfo{pages}{3675--3680}.
\newblock


\bibitem[Kingma and Ba(2014)]%
        {kingma2014adam}
\bibfield{author}{\bibinfo{person}{Diederik~P Kingma} {and} \bibinfo{person}{Jimmy Ba}.} \bibinfo{year}{2014}\natexlab{}.
\newblock \showarticletitle{Adam: A method for stochastic optimization}.
\newblock \bibinfo{journal}{\emph{arXiv preprint arXiv:1412.6980}} (\bibinfo{year}{2014}).
\newblock


\bibitem[Ko{\c{c}}er and Arslan(2010)]%
        {koccer2010genetic}
\bibfield{author}{\bibinfo{person}{Bar{\i}{\c{s}} Ko{\c{c}}er} {and} \bibinfo{person}{Ahmet Arslan}.} \bibinfo{year}{2010}\natexlab{}.
\newblock \showarticletitle{Genetic transfer learning}.
\newblock \bibinfo{journal}{\emph{Expert Systems with Applications}} \bibinfo{volume}{37}, \bibinfo{number}{10} (\bibinfo{year}{2010}), \bibinfo{pages}{6997--7002}.
\newblock


\bibitem[Kulsoom et~al\mbox{.}(2022)]%
        {kulsoom2022review}
\bibfield{author}{\bibinfo{person}{Farzana Kulsoom}, \bibinfo{person}{Sanam Narejo}, \bibinfo{person}{Zahid Mehmood}, \bibinfo{person}{Hassan~Nazeer Chaudhry}, \bibinfo{person}{Ayesha Butt}, {and} \bibinfo{person}{Ali~Kashif Bashir}.} \bibinfo{year}{2022}\natexlab{}.
\newblock \showarticletitle{A review of machine learning-based human activity recognition for diverse applications}.
\newblock \bibinfo{journal}{\emph{Neural Computing and Applications}} \bibinfo{volume}{34}, \bibinfo{number}{21} (\bibinfo{year}{2022}), \bibinfo{pages}{18289--18324}.
\newblock


\bibitem[Li et~al\mbox{.}(2019)]%
        {li2019relation}
\bibfield{author}{\bibinfo{person}{Linjie Li}, \bibinfo{person}{Zhe Gan}, \bibinfo{person}{Yu Cheng}, {and} \bibinfo{person}{Jingjing Liu}.} \bibinfo{year}{2019}\natexlab{}.
\newblock \showarticletitle{Relation-aware graph attention network for visual question answering}. In \bibinfo{booktitle}{\emph{Proceedings of the IEEE/CVF international conference on computer vision}}. \bibinfo{pages}{10313--10322}.
\newblock


\bibitem[Li et~al\mbox{.}(2024)]%
        {li2024synthetic}
\bibfield{author}{\bibinfo{person}{Yuhang Li}, \bibinfo{person}{Xin Dong}, \bibinfo{person}{Chen Chen}, \bibinfo{person}{Jingtao Li}, \bibinfo{person}{Yuxin Wen}, \bibinfo{person}{Michael Spranger}, {and} \bibinfo{person}{Lingjuan Lyu}.} \bibinfo{year}{2024}\natexlab{}.
\newblock \showarticletitle{Is Synthetic Image Useful for Transfer Learning? An Investigation into Data Generation, Volume, and Utilization}.
\newblock \bibinfo{journal}{\emph{arXiv preprint arXiv:2403.19866}} (\bibinfo{year}{2024}).
\newblock


\bibitem[Liciotti et~al\mbox{.}(2020)]%
        {liciotti2020sequential}
\bibfield{author}{\bibinfo{person}{Daniele Liciotti}, \bibinfo{person}{Michele Bernardini}, \bibinfo{person}{Luca Romeo}, {and} \bibinfo{person}{Emanuele Frontoni}.} \bibinfo{year}{2020}\natexlab{}.
\newblock \showarticletitle{A sequential deep learning application for recognising human activities in smart homes}.
\newblock \bibinfo{journal}{\emph{Neurocomputing}}  \bibinfo{volume}{396} (\bibinfo{year}{2020}), \bibinfo{pages}{501--513}.
\newblock


\bibitem[Liu et~al\mbox{.}(2024)]%
        {liu2024visual}
\bibfield{author}{\bibinfo{person}{Haotian Liu}, \bibinfo{person}{Chunyuan Li}, \bibinfo{person}{Qingyang Wu}, {and} \bibinfo{person}{Yong~Jae Lee}.} \bibinfo{year}{2024}\natexlab{}.
\newblock \showarticletitle{Visual instruction tuning}.
\newblock \bibinfo{journal}{\emph{Advances in neural information processing systems}}  \bibinfo{volume}{36} (\bibinfo{year}{2024}).
\newblock


\bibitem[Liu et~al\mbox{.}(2019)]%
        {liu2019roberta}
\bibfield{author}{\bibinfo{person}{Yinhan Liu}, \bibinfo{person}{Myle Ott}, \bibinfo{person}{Naman Goyal}, \bibinfo{person}{Jingfei Du}, \bibinfo{person}{Mandar Joshi}, \bibinfo{person}{Danqi Chen}, \bibinfo{person}{Omer Levy}, \bibinfo{person}{Mike Lewis}, \bibinfo{person}{Luke Zettlemoyer}, {and} \bibinfo{person}{Veselin Stoyanov}.} \bibinfo{year}{2019}\natexlab{}.
\newblock \showarticletitle{Roberta: A robustly optimized bert pretraining approach}.
\newblock \bibinfo{journal}{\emph{arXiv preprint arXiv:1907.11692}} (\bibinfo{year}{2019}).
\newblock


\bibitem[Maurer et~al\mbox{.}(2013)]%
        {maurer2013sparse}
\bibfield{author}{\bibinfo{person}{Andreas Maurer}, \bibinfo{person}{Massi Pontil}, {and} \bibinfo{person}{Bernardino Romera-Paredes}.} \bibinfo{year}{2013}\natexlab{}.
\newblock \showarticletitle{Sparse coding for multitask and transfer learning}. In \bibinfo{booktitle}{\emph{International conference on machine learning}}. PMLR, \bibinfo{pages}{343--351}.
\newblock


\bibitem[Morita et~al\mbox{.}(2023)]%
        {morita2023health}
\bibfield{author}{\bibinfo{person}{Plinio~P Morita}, \bibinfo{person}{Kirti~Sundar Sahu}, {and} \bibinfo{person}{Arlene Oetomo}.} \bibinfo{year}{2023}\natexlab{}.
\newblock \showarticletitle{Health monitoring using smart home technologies: Scoping review}.
\newblock \bibinfo{journal}{\emph{JMIR mHealth and uHealth}}  \bibinfo{volume}{11} (\bibinfo{year}{2023}), \bibinfo{pages}{e37347}.
\newblock


\bibitem[Ni et~al\mbox{.}(2021)]%
        {ni2021sentence}
\bibfield{author}{\bibinfo{person}{Jianmo Ni}, \bibinfo{person}{Gustavo~Hernandez Abrego}, \bibinfo{person}{Noah Constant}, \bibinfo{person}{Ji Ma}, \bibinfo{person}{Keith~B Hall}, \bibinfo{person}{Daniel Cer}, {and} \bibinfo{person}{Yinfei Yang}.} \bibinfo{year}{2021}\natexlab{}.
\newblock \showarticletitle{Sentence-t5: Scalable sentence encoders from pre-trained text-to-text models}.
\newblock \bibinfo{journal}{\emph{arXiv preprint arXiv:2108.08877}} (\bibinfo{year}{2021}).
\newblock


\bibitem[Ord{\'o}{\~n}ez and Roggen(2016)]%
        {ordonez2016deep}
\bibfield{author}{\bibinfo{person}{Francisco~Javier Ord{\'o}{\~n}ez} {and} \bibinfo{person}{Daniel Roggen}.} \bibinfo{year}{2016}\natexlab{}.
\newblock \showarticletitle{Deep convolutional and lstm recurrent neural networks for multimodal wearable activity recognition}.
\newblock \bibinfo{journal}{\emph{Sensors}} \bibinfo{volume}{16}, \bibinfo{number}{1} (\bibinfo{year}{2016}), \bibinfo{pages}{115}.
\newblock


\bibitem[Pan and Yang(2009)]%
        {pan2009survey}
\bibfield{author}{\bibinfo{person}{Sinno~Jialin Pan} {and} \bibinfo{person}{Qiang Yang}.} \bibinfo{year}{2009}\natexlab{}.
\newblock \showarticletitle{A survey on transfer learning}.
\newblock \bibinfo{journal}{\emph{IEEE Transactions on knowledge and data engineering}} \bibinfo{volume}{22}, \bibinfo{number}{10} (\bibinfo{year}{2009}), \bibinfo{pages}{1345--1359}.
\newblock


\bibitem[Parnami and Lee(2022)]%
        {parnami2022learning}
\bibfield{author}{\bibinfo{person}{Archit Parnami} {and} \bibinfo{person}{Minwoo Lee}.} \bibinfo{year}{2022}\natexlab{}.
\newblock \showarticletitle{Learning from few examples: A summary of approaches to few-shot learning}.
\newblock \bibinfo{journal}{\emph{arXiv preprint arXiv:2203.04291}} (\bibinfo{year}{2022}).
\newblock


\bibitem[Paszke et~al\mbox{.}(2019)]%
        {paszke2019pytorch}
\bibfield{author}{\bibinfo{person}{Adam Paszke}, \bibinfo{person}{Sam Gross}, \bibinfo{person}{Francisco Massa}, \bibinfo{person}{Adam Lerer}, \bibinfo{person}{James Bradbury}, \bibinfo{person}{Gregory Chanan}, \bibinfo{person}{Trevor Killeen}, \bibinfo{person}{Zeming Lin}, \bibinfo{person}{Natalia Gimelshein}, \bibinfo{person}{Luca Antiga}, {et~al\mbox{.}}} \bibinfo{year}{2019}\natexlab{}.
\newblock \showarticletitle{Pytorch: An imperative style, high-performance deep learning library}.
\newblock \bibinfo{journal}{\emph{Advances in neural information processing systems}}  \bibinfo{volume}{32} (\bibinfo{year}{2019}).
\newblock


\bibitem[Qin et~al\mbox{.}(2019)]%
        {qin2019cross}
\bibfield{author}{\bibinfo{person}{Xin Qin}, \bibinfo{person}{Yiqiang Chen}, \bibinfo{person}{Jindong Wang}, {and} \bibinfo{person}{Chaohui Yu}.} \bibinfo{year}{2019}\natexlab{}.
\newblock \showarticletitle{Cross-dataset activity recognition via adaptive spatial-temporal transfer learning}.
\newblock \bibinfo{journal}{\emph{Proceedings of the ACM on Interactive, Mobile, Wearable and Ubiquitous Technologies}} \bibinfo{volume}{3}, \bibinfo{number}{4} (\bibinfo{year}{2019}), \bibinfo{pages}{1--25}.
\newblock


\bibitem[Quigley et~al\mbox{.}(2018)]%
        {quigley2018comparative}
\bibfield{author}{\bibinfo{person}{Bronagh Quigley}, \bibinfo{person}{Mark Donnelly}, \bibinfo{person}{George Moore}, {and} \bibinfo{person}{Leo Galway}.} \bibinfo{year}{2018}\natexlab{}.
\newblock \showarticletitle{A comparative analysis of windowing approaches in dense sensing environments}. In \bibinfo{booktitle}{\emph{Proceedings}}, Vol.~\bibinfo{volume}{2}. MDPI, \bibinfo{pages}{1245}.
\newblock


\bibitem[Raina et~al\mbox{.}(2007)]%
        {raina2007self}
\bibfield{author}{\bibinfo{person}{Rajat Raina}, \bibinfo{person}{Alexis Battle}, \bibinfo{person}{Honglak Lee}, \bibinfo{person}{Benjamin Packer}, {and} \bibinfo{person}{Andrew~Y Ng}.} \bibinfo{year}{2007}\natexlab{}.
\newblock \showarticletitle{Self-taught learning: transfer learning from unlabeled data}. In \bibinfo{booktitle}{\emph{Proceedings of the 24th international conference on Machine learning}}. \bibinfo{pages}{759--766}.
\newblock


\bibitem[Rajani et~al\mbox{.}(2019)]%
        {rajani2019explain}
\bibfield{author}{\bibinfo{person}{Nazneen~Fatema Rajani}, \bibinfo{person}{Bryan McCann}, \bibinfo{person}{Caiming Xiong}, {and} \bibinfo{person}{Richard Socher}.} \bibinfo{year}{2019}\natexlab{}.
\newblock \showarticletitle{Explain yourself! leveraging language models for commonsense reasoning}.
\newblock \bibinfo{journal}{\emph{arXiv preprint arXiv:1906.02361}} (\bibinfo{year}{2019}).
\newblock


\bibitem[Rashidi and Cook(2009)]%
        {rashidi2009keeping}
\bibfield{author}{\bibinfo{person}{Parisa Rashidi} {and} \bibinfo{person}{Diane~J Cook}.} \bibinfo{year}{2009}\natexlab{}.
\newblock \showarticletitle{Keeping the resident in the loop: Adapting the smart home to the user}.
\newblock \bibinfo{journal}{\emph{IEEE Transactions on systems, man, and cybernetics-part A: systems and humans}} \bibinfo{volume}{39}, \bibinfo{number}{5} (\bibinfo{year}{2009}), \bibinfo{pages}{949--959}.
\newblock


\bibitem[Rashidi and Cook(2011)]%
        {rashidi2011domain}
\bibfield{author}{\bibinfo{person}{Parisa Rashidi} {and} \bibinfo{person}{Diane~J Cook}.} \bibinfo{year}{2011}\natexlab{}.
\newblock \showarticletitle{Domain selection and adaptation in smart homes}. In \bibinfo{booktitle}{\emph{International Conference on Smart Homes and Health Telematics}}. Springer, \bibinfo{pages}{17--24}.
\newblock


\bibitem[Reimers and Gurevych(2019)]%
        {reimers2019sentence}
\bibfield{author}{\bibinfo{person}{Nils Reimers} {and} \bibinfo{person}{Iryna Gurevych}.} \bibinfo{year}{2019}\natexlab{}.
\newblock \showarticletitle{Sentence-bert: Sentence embeddings using siamese bert-networks}.
\newblock \bibinfo{journal}{\emph{arXiv preprint arXiv:1908.10084}} (\bibinfo{year}{2019}).
\newblock


\bibitem[Rotem et~al\mbox{.}(2022)]%
        {rotem2022transfer}
\bibfield{author}{\bibinfo{person}{Yarden Rotem}, \bibinfo{person}{Nathaniel Shimoni}, \bibinfo{person}{Lior Rokach}, {and} \bibinfo{person}{Bracha Shapira}.} \bibinfo{year}{2022}\natexlab{}.
\newblock \showarticletitle{Transfer learning for time series classification using synthetic data generation}. In \bibinfo{booktitle}{\emph{International Symposium on Cyber Security, Cryptology, and Machine Learning}}. Springer, \bibinfo{pages}{232--246}.
\newblock


\bibitem[Sanabria et~al\mbox{.}(2021)]%
        {sanabria2021unsupervised}
\bibfield{author}{\bibinfo{person}{Andrea~Rosales Sanabria}, \bibinfo{person}{Franco Zambonelli}, {and} \bibinfo{person}{Juan Ye}.} \bibinfo{year}{2021}\natexlab{}.
\newblock \showarticletitle{Unsupervised domain adaptation in activity recognition: A GAN-based approach}.
\newblock \bibinfo{journal}{\emph{IEEE Access}}  \bibinfo{volume}{9} (\bibinfo{year}{2021}), \bibinfo{pages}{19421--19438}.
\newblock


\bibitem[Sanh et~al\mbox{.}(2019)]%
        {Sanh2019DistilBERTAD}
\bibfield{author}{\bibinfo{person}{Victor Sanh}, \bibinfo{person}{Lysandre Debut}, \bibinfo{person}{Julien Chaumond}, {and} \bibinfo{person}{Thomas Wolf}.} \bibinfo{year}{2019}\natexlab{}.
\newblock \showarticletitle{DistilBERT, a distilled version of BERT: smaller, faster, cheaper and lighter}.
\newblock \bibinfo{journal}{\emph{ArXiv}}  \bibinfo{volume}{abs/1910.01108} (\bibinfo{year}{2019}).
\newblock


\bibitem[Sedky et~al\mbox{.}(2018)]%
        {sedky2018evaluating}
\bibfield{author}{\bibinfo{person}{Mohamed Sedky}, \bibinfo{person}{Christopher Howard}, \bibinfo{person}{Talal Alshammari}, {and} \bibinfo{person}{Nasser Alshammari}.} \bibinfo{year}{2018}\natexlab{}.
\newblock \showarticletitle{Evaluating machine learning techniques for activity classification in smart home environments}.
\newblock \bibinfo{journal}{\emph{International Journal of Information Systems and Computer Sciences}} \bibinfo{volume}{12}, \bibinfo{number}{2} (\bibinfo{year}{2018}), \bibinfo{pages}{48--54}.
\newblock


\bibitem[Settles(2009)]%
        {settles2009active}
\bibfield{author}{\bibinfo{person}{Burr Settles}.} \bibinfo{year}{2009}\natexlab{}.
\newblock \showarticletitle{Active learning literature survey}.
\newblock  (\bibinfo{year}{2009}).
\newblock


\bibitem[Singh et~al\mbox{.}(2017)]%
        {singh2017convolutional}
\bibfield{author}{\bibinfo{person}{Deepika Singh}, \bibinfo{person}{Erinc Merdivan}, \bibinfo{person}{Sten Hanke}, \bibinfo{person}{Johannes Kropf}, \bibinfo{person}{Matthieu Geist}, {and} \bibinfo{person}{Andreas Holzinger}.} \bibinfo{year}{2017}\natexlab{}.
\newblock \showarticletitle{Convolutional and recurrent neural networks for activity recognition in smart environment}. In \bibinfo{booktitle}{\emph{Towards Integrative Machine Learning and Knowledge Extraction: BIRS Workshop, Banff, AB, Canada, July 24-26, 2015, Revised Selected Papers}}. Springer, \bibinfo{pages}{194--205}.
\newblock


\bibitem[Snell et~al\mbox{.}(2017)]%
        {snell2017prototypical}
\bibfield{author}{\bibinfo{person}{Jake Snell}, \bibinfo{person}{Kevin Swersky}, {and} \bibinfo{person}{Richard Zemel}.} \bibinfo{year}{2017}\natexlab{}.
\newblock \showarticletitle{Prototypical networks for few-shot learning}.
\newblock \bibinfo{journal}{\emph{Advances in neural information processing systems}}  \bibinfo{volume}{30} (\bibinfo{year}{2017}).
\newblock


\bibitem[Sun et~al\mbox{.}(2019)]%
        {sun2019meta}
\bibfield{author}{\bibinfo{person}{Qianru Sun}, \bibinfo{person}{Yaoyao Liu}, \bibinfo{person}{Tat-Seng Chua}, {and} \bibinfo{person}{Bernt Schiele}.} \bibinfo{year}{2019}\natexlab{}.
\newblock \showarticletitle{Meta-transfer learning for few-shot learning}. In \bibinfo{booktitle}{\emph{Proceedings of the IEEE/CVF conference on computer vision and pattern recognition}}. \bibinfo{pages}{403--412}.
\newblock


\bibitem[Sung et~al\mbox{.}(2018)]%
        {sung2018learning}
\bibfield{author}{\bibinfo{person}{Flood Sung}, \bibinfo{person}{Yongxin Yang}, \bibinfo{person}{Li Zhang}, \bibinfo{person}{Tao Xiang}, \bibinfo{person}{Philip~HS Torr}, {and} \bibinfo{person}{Timothy~M Hospedales}.} \bibinfo{year}{2018}\natexlab{}.
\newblock \showarticletitle{Learning to compare: Relation network for few-shot learning}. In \bibinfo{booktitle}{\emph{Proceedings of the IEEE conference on computer vision and pattern recognition}}. \bibinfo{pages}{1199--1208}.
\newblock


\bibitem[Torrey and Shavlik(2010)]%
        {torrey2010transfer}
\bibfield{author}{\bibinfo{person}{Lisa Torrey} {and} \bibinfo{person}{Jude Shavlik}.} \bibinfo{year}{2010}\natexlab{}.
\newblock \showarticletitle{Transfer learning}.
\newblock In \bibinfo{booktitle}{\emph{Handbook of research on machine learning applications and trends: algorithms, methods, and techniques}}. \bibinfo{publisher}{IGI global}, \bibinfo{pages}{242--264}.
\newblock


\bibitem[Wang et~al\mbox{.}(2019)]%
        {wang2019instance}
\bibfield{author}{\bibinfo{person}{Tianyang Wang}, \bibinfo{person}{Jun Huan}, {and} \bibinfo{person}{Michelle Zhu}.} \bibinfo{year}{2019}\natexlab{}.
\newblock \showarticletitle{Instance-based deep transfer learning}. In \bibinfo{booktitle}{\emph{2019 IEEE Winter Conference on Applications of Computer Vision (WACV)}}. IEEE, \bibinfo{pages}{367--375}.
\newblock


\bibitem[Wang et~al\mbox{.}(2020)]%
        {wang2020generalizing}
\bibfield{author}{\bibinfo{person}{Yaqing Wang}, \bibinfo{person}{Quanming Yao}, \bibinfo{person}{James~T Kwok}, {and} \bibinfo{person}{Lionel~M Ni}.} \bibinfo{year}{2020}\natexlab{}.
\newblock \showarticletitle{Generalizing from a few examples: A survey on few-shot learning}.
\newblock \bibinfo{journal}{\emph{ACM computing surveys (csur)}} \bibinfo{volume}{53}, \bibinfo{number}{3} (\bibinfo{year}{2020}), \bibinfo{pages}{1--34}.
\newblock


\bibitem[Weiss et~al\mbox{.}(2016)]%
        {weiss2016survey}
\bibfield{author}{\bibinfo{person}{Karl Weiss}, \bibinfo{person}{Taghi~M Khoshgoftaar}, {and} \bibinfo{person}{DingDing Wang}.} \bibinfo{year}{2016}\natexlab{}.
\newblock \showarticletitle{A survey of transfer learning}.
\newblock \bibinfo{journal}{\emph{Journal of Big data}}  \bibinfo{volume}{3} (\bibinfo{year}{2016}), \bibinfo{pages}{1--40}.
\newblock


\bibitem[Wolf et~al\mbox{.}(2019)]%
        {wolf2019huggingface}
\bibfield{author}{\bibinfo{person}{Thomas Wolf}, \bibinfo{person}{Lysandre Debut}, \bibinfo{person}{Victor Sanh}, \bibinfo{person}{Julien Chaumond}, \bibinfo{person}{Clement Delangue}, \bibinfo{person}{Anthony Moi}, \bibinfo{person}{Pierric Cistac}, \bibinfo{person}{Tim Rault}, \bibinfo{person}{R{\'e}mi Louf}, \bibinfo{person}{Morgan Funtowicz}, {et~al\mbox{.}}} \bibinfo{year}{2019}\natexlab{}.
\newblock \showarticletitle{Huggingface's transformers: State-of-the-art natural language processing}.
\newblock \bibinfo{journal}{\emph{arXiv preprint arXiv:1910.03771}} (\bibinfo{year}{2019}).
\newblock


\bibitem[Wu et~al\mbox{.}(2022)]%
        {wu2022cluster}
\bibfield{author}{\bibinfo{person}{Renzhi Wu}, \bibinfo{person}{Nilaksh Das}, \bibinfo{person}{Sanya Chaba}, \bibinfo{person}{Sakshi Gandhi}, \bibinfo{person}{Duen~Horng Chau}, {and} \bibinfo{person}{Xu Chu}.} \bibinfo{year}{2022}\natexlab{}.
\newblock \showarticletitle{A cluster-then-label approach for few-shot learning with application to automatic image data labeling}.
\newblock \bibinfo{journal}{\emph{ACM Journal of Data and Information Quality (JDIQ)}} \bibinfo{volume}{14}, \bibinfo{number}{3} (\bibinfo{year}{2022}), \bibinfo{pages}{1--23}.
\newblock


\bibitem[Zhai et~al\mbox{.}(2019)]%
        {zhai2019s4l}
\bibfield{author}{\bibinfo{person}{Xiaohua Zhai}, \bibinfo{person}{Avital Oliver}, \bibinfo{person}{Alexander Kolesnikov}, {and} \bibinfo{person}{Lucas Beyer}.} \bibinfo{year}{2019}\natexlab{}.
\newblock \showarticletitle{S4l: Self-supervised semi-supervised learning}. In \bibinfo{booktitle}{\emph{Proceedings of the IEEE/CVF international conference on computer vision}}. \bibinfo{pages}{1476--1485}.
\newblock


\bibitem[Zhao et~al\mbox{.}(2020)]%
        {zhao2020makes}
\bibfield{author}{\bibinfo{person}{Nanxuan Zhao}, \bibinfo{person}{Zhirong Wu}, \bibinfo{person}{Rynson~WH Lau}, {and} \bibinfo{person}{Stephen Lin}.} \bibinfo{year}{2020}\natexlab{}.
\newblock \showarticletitle{What makes instance discrimination good for transfer learning?}
\newblock \bibinfo{journal}{\emph{arXiv preprint arXiv:2006.06606}} (\bibinfo{year}{2020}).
\newblock


\bibitem[Zhou et~al\mbox{.}(2020a)]%
        {zhou2020graph}
\bibfield{author}{\bibinfo{person}{Jie Zhou}, \bibinfo{person}{Ganqu Cui}, \bibinfo{person}{Shengding Hu}, \bibinfo{person}{Zhengyan Zhang}, \bibinfo{person}{Cheng Yang}, \bibinfo{person}{Zhiyuan Liu}, \bibinfo{person}{Lifeng Wang}, \bibinfo{person}{Changcheng Li}, {and} \bibinfo{person}{Maosong Sun}.} \bibinfo{year}{2020}\natexlab{a}.
\newblock \showarticletitle{Graph neural networks: A review of methods and applications}.
\newblock \bibinfo{journal}{\emph{AI open}}  \bibinfo{volume}{1} (\bibinfo{year}{2020}), \bibinfo{pages}{57--81}.
\newblock


\bibitem[Zhou et~al\mbox{.}(2020b)]%
        {zhou2020evaluating}
\bibfield{author}{\bibinfo{person}{Xuhui Zhou}, \bibinfo{person}{Yue Zhang}, \bibinfo{person}{Leyang Cui}, {and} \bibinfo{person}{Dandan Huang}.} \bibinfo{year}{2020}\natexlab{b}.
\newblock \showarticletitle{Evaluating commonsense in pre-trained language models}. In \bibinfo{booktitle}{\emph{Proceedings of the AAAI conference on artificial intelligence}}, Vol.~\bibinfo{volume}{34}. \bibinfo{pages}{9733--9740}.
\newblock


\bibitem[Zhuang et~al\mbox{.}(2020)]%
        {zhuang2020comprehensive}
\bibfield{author}{\bibinfo{person}{Fuzhen Zhuang}, \bibinfo{person}{Zhiyuan Qi}, \bibinfo{person}{Keyu Duan}, \bibinfo{person}{Dongbo Xi}, \bibinfo{person}{Yongchun Zhu}, \bibinfo{person}{Hengshu Zhu}, \bibinfo{person}{Hui Xiong}, {and} \bibinfo{person}{Qing He}.} \bibinfo{year}{2020}\natexlab{}.
\newblock \showarticletitle{A comprehensive survey on transfer learning}.
\newblock \bibinfo{journal}{\emph{Proc. IEEE}} \bibinfo{volume}{109}, \bibinfo{number}{1} (\bibinfo{year}{2020}), \bibinfo{pages}{43--76}.
\newblock


\end{thebibliography}

\clearpage
\appendix
\section{Appendix}
\subsection{Mapping of Activities}
\label{sec:appendix:activity_mapping}
While performing layout-agnostic HAR, we first convert the set of activities in each smart home into a set of common activities, based on their semantic similarity.
The mapping scheme used to perform this conversion is provided in Table \ref{tab:activity_translation}.
In particular, we map each activity in every smart home into one of the following common activities: $\{$Relax, Cook, Leave Home, Enter Home, Sleep, Eat, Work, Bed to Toilet, Bathing, Take Medicine, Personal Hygiene, and Other$\}$. 
Once this conversion is performed, we consider the common activities present in both the source and the target smart homes while performing layout-agnostic HAR.  
\begin{table*}[t!]
	\centering
	\small
	\begin{tabular}{P{3.7cm} | P{3.7cm} | P{3.7cm} | P{3.7cm}}
        \toprule
		Aruba & Milan & Kyoto7 & Cairo \\ 
		\midrule
		Relax $\rightarrow$ Relax & Kitchen Activity $\rightarrow$ Cook & Meal Preparation $\rightarrow$ Cook & Leave Home $\rightarrow$ Leave Home \\ 
            Meal Preparation $\rightarrow$ Cook & Guest Bathroom $\rightarrow$ Bathing & R1 Work $\rightarrow$ Work & Night Wandering $\rightarrow$ Other \\ 
            Enter Home $\rightarrow$ Enter Home& Read $\rightarrow$ Relax & R1 Personal Hygiene $\rightarrow$ Personal Hygiene  & R1 Wake $\rightarrow$ Other \\ 
            Leave Home $\rightarrow$ Leave Home & Master Bathroom $\rightarrow$ Bathing & R2 Work $\rightarrow$ Work & R2 Wake $\rightarrow$ Other \\
            Sleeping $\rightarrow$ Sleep & Leave Home $\rightarrow$ Leave Home & R2 Bed to Toilet $\rightarrow$ Bed to Toilet & R1 Sleep $\rightarrow$ Sleep \\
            Eating $\rightarrow$ Eat & Master Bedroom Activity $\rightarrow$ Other & R2 Personal Hygiene $\rightarrow$ Personal Hygiene & R2 Sleep $\rightarrow$ Sleep \\
            Work $\rightarrow$ Work & Watch TV $\rightarrow$ Relax & R1 Sleep $\rightarrow$ Sleep & Breakfast $\rightarrow$ Eat \\
            Bed to Toilet $\rightarrow$ Bed to Toilet& Sleep $\rightarrow$ Sleep & R2 Sleep $\rightarrow$ Sleep & R1 Work in Office $\rightarrow$ Work\\
            Wash Dishes $\rightarrow$ Work& Bed to Toilet $\rightarrow$ Bed to Toilet & R1 Bed to Toilet $\rightarrow$ Bed to Toilet & R2 Take Medicine $\rightarrow$ Take Medicine\\
            Housekeeping $\rightarrow$ Work& Desk Activity $\rightarrow$ Work & Watch TV $\rightarrow$ Relax & Dinner $\rightarrow$ Eat \\
            Resperate $\rightarrow$ Other& Morning Meds $\rightarrow$ Take Medicine & Study $\rightarrow$ Other & Lunch $\rightarrow$ Eat \\
            Other $\rightarrow$ Other& Chores $\rightarrow$ Work & Clean $\rightarrow$ Work & Bed to Toilet $\rightarrow$ Bed to Toilet \\
            & Dining Room Activity $\rightarrow$ Eat & Wash Bathtub $\rightarrow$ Other & Laundry $\rightarrow$ Work \\
            & Evening Meds $\rightarrow$ Take Medicine & Other $\rightarrow$ Other & Other $\rightarrow$ Other \\
            & Meditate $\rightarrow$ Other & & \\
            & Other $\rightarrow$ Other & & \\
		\bottomrule
	\end{tabular}
 \caption{Mapping of activities into the common activity set 
 }
	\label{tab:activity_translation}
\end{table*}

\subsection{Converting floorplans and sensor metadata to text descriptions in different TDOST versions}

We have described the templates for incorporating varying contexts in different TDOST versions in Section \ref{TDOST-Variants}.
Here, we provide the conversion maps for constructing TDOST from sensor layout for some of the publicly available CASAS datasets used in our work, namely Aruba, Milan, and Cairo. 
We visually inspect the sensor layout images released publicly along with the CASAS datasets to encode the symbolic position of each sensor in the smart home. 
Similarly, the sensor characteristics such as the number of sensors and their types are encoded from publicly released dataset metadata and accompanying documents. 


\subsubsection{Conversion map for TDOST-Basic:}\label{appendix_tdost_basic}
\begin{figure}[]
  \centering
  \includegraphics[width=0.9\linewidth]{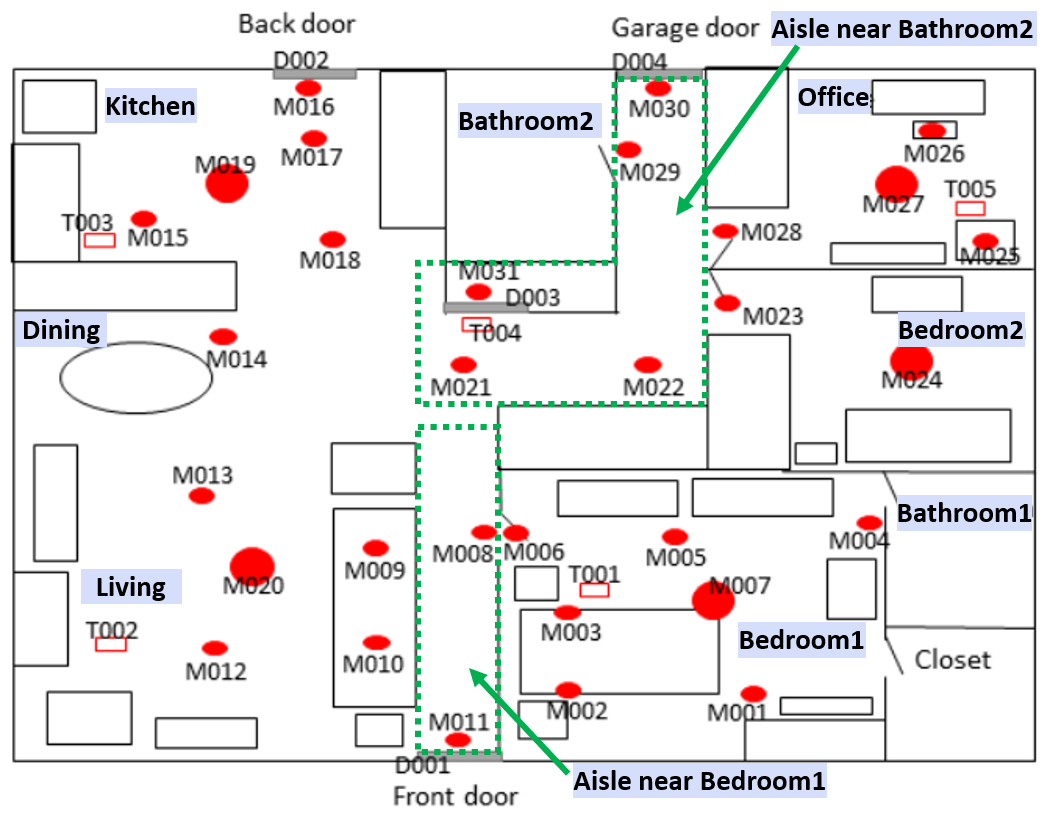} 
  \caption{Aruba sensor layout and floorplan (taken with permission from \cite{hiremath2022bootstrapping} and \cite{cook2012casas}) }

\label{fig:aruba_floorplan}
\end{figure}

In Tab. \ref{tab:aruba_tdost_basic}, we give examples of mapping raw sensor triggers to textual descriptions for CASAS-Aruba dataset using the template for TDOST-Basic variant.
We picked Aruba as a representative example dataset, used its floorplan shown in Fig. \ref{fig:aruba_floorplan} to demonstrate the procedure for building TDOST-Basic.
A similar conversion map can be built for the rest of the datasets by incorporating metadata from the publicly available repositories.

\begin{table}
\centering
\begin{adjustbox}{totalheight=\textheight}
\begin{tabularx}{\linewidth}{p{3cm}p{4cm}p{1.5cm}p{1.5cm}X}
\toprule
\textbf{Sensor Id} & \textbf{Location Context} & \textbf{Type} & \textbf{Value} & \textbf{TDOST-Basic Example}\\
\midrule
M001, M002, M003, M005, M007 & In first Bedroom & Motion & OFF & \textit{Motion sensor in first bedroom fired with value OFF}\\
M013 & Between dining area and living room & Motion & ON & \textit{Motion sensor between dining area and living room fired with value ON}\\
M009, M010 & Between living room and home entrance aisle & Motion & ON & \textit{Motion sensor between living room and home entrance aisle fired with value ON}\\
M011, M008 & In home entrance aisle & Motion & ON & \textit{Motion sensor in home entrance aisle fired with value ON}\\
M004 & Between the first bedroom and first bathroom & Motion & ON & \textit{Motion sensor between the first bedroom and first bathroom fired with value ON}\\
M022 & In aisle between second bathroom and second bedroom & Motion & ON & \textit{Motion sensor in the aisle between second bathroom and second bedroom fired with value ON}\\
M024, M023 & In second bedroom & Motion & OFF & \textit{Motion sensor in second bedroom fired with value OFF}\\
M029 & Between aisle and second bathroom & Motion & ON & \textit{Motion sensor between aisle and second bathroom fired with value ON}\\
M030 & In aisle between garage door and second bathroom & Motion & OFF & \textit{Motion sensor in the aisle between garage door and second bathroom fired with value OFF}\\
M028 & Between office and garage door aisle & Motion & ON & \textit{Motion sensor between office and garage door aisle fired with value ON}\\
M027, M026, M025 & In office & Motion & OFF & \textit{Motion sensor in office fired with value OFF}\\
M015, M019, M017, M016 & In Kitchen & Motion & ON & \textit{Motion sensors in kitchen fired with value ON/OFF}\\
D002 & Between kitchen and back door & Door & OPEN & \textit{Door sensor between kitchen and back door fired with value OPEN}\\
M012, M020 & In Living room & Motion & OFF & \textit{Motion sensors in living room fired with value OFF}\\
D001 & In home entrance aisle & Door & CLOSE & \textit{Door sensor in home entrance aisle fired with value CLOSE}\\
T001 & In the first bedroom & Temperature & 25 & \textit{Temperature sensor in first bedroom fired with value twenty-five}\\
T004 & In aisle between second bathroom and dining area & Temperature & 22 & \textit{Temperature sensor in aisle between second bathroom and dining area fired with value twenty-two}\\
D004 & On garage door & Door & OPEN & \textit{Door sensor on garage door fired with value OPEN}\\
\bottomrule
\end{tabularx}
\end{adjustbox}
\caption{ Conversion table to build textual description using TDOST-Basic variant for a subset of sensors in CASAS-Aruba dataset. } 
\label{tab:aruba_tdost_basic}

\end{table}

\subsubsection{Conversion map for TDOST-Temporal:}\label{appendix_tdost_temporal}
\begin{figure}[]
  \centering
  \includegraphics[width=0.9\linewidth]{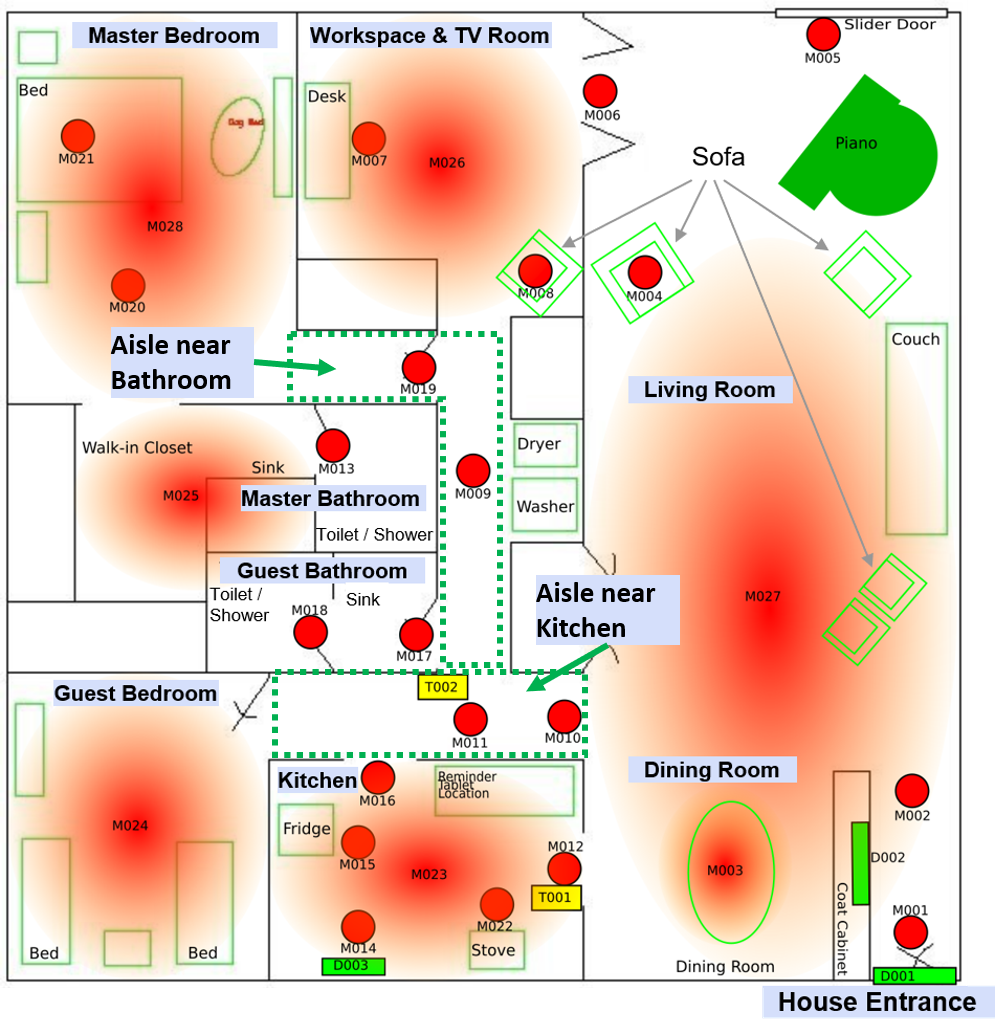} 
  \caption{Milan sensor layout and floorplan (taken with permission from \cite{hiremath2022bootstrapping} and \cite{cook2012casas}) }

\label{fig:milan_floorplan}
\end{figure}

For showing the conversion map of TDOST-Temporal, we chose the CASAS-Milan sensor layout shown in Fig.\ref{fig:milan_floorplan}. 
We provide details of the granular sensor location map, relative temporal information, and conversion process used for building the  TDOST-Temporal in Tab. \ref{tab:milan_tdost_temporal}.
An arbitrary time lag between the given example and the previous sensor trigger is assumed and added to the examples shown here.
For some of the examples, we add an absolute timestamp replicating the setup when the sensor trigger is the first reading in a given activity sequence.
For each sensor reading instance, we assume a random value (specified in the Value column in Tab.\ref{tab:milan_tdost_temporal} ) and show its usage in the final constructed sentence.

\begin{table}
\centering
\begin{adjustbox}{totalheight=\textheight}
\begin{tabularx}{\linewidth}{p{3cm}p{4cm}p{1.5cm}p{1.5cm}X}
\toprule
\textbf{Sensor Id} & \textbf{Location Context} & \textbf{Type} & \textbf{Value} & \textbf{TDOST-Temporal Example}\\
\midrule
M001 & near home entrance & Motion & OFF & \textit{After seven seconds, motion sensor near home entrance fired with value OFF} \\
M002 & near home entrance towards living room & Motion & ON & \textit{After two seconds, motion sensor near home entrance towards living room fired with value ON} \\
M005 & in living room near slider door & Motion & OFF & \textit{After thirty-five seconds, Motion sensor in living room near slider door fired with value OFF} \\
M006 & between living room and workspace / TV room & Motion & ON & \textit{After four seconds, Motion sensor between living room and workspace/TV room fired with value ON} \\
M007 & in workspace / TV room near desk & Motion & ON & \textit{After sixty-four seconds, Motion sensor in workspace/TV room near desk fired with value ON} \\
M010 & in corridor between dining room and kitchen & Motion & ON & \textit{After sixty-four seconds, Motion sensor in corridor between dining room and kitchen fired with value ON} \\

M012 & between dining room and kitchen & Motion & OFF & \textit{Motion sensor between dining room and kitchen fired with value OFF at twelve hours} \\
M013 & in bathroom near sink & Motion & ON & \textit{After one seconds, Motion sensor in bathroom near sink fired with value ON} \\
M014 & in kitchen near door & Motion & OFF & \textit{After forty seconds, Motion sensor in kitchen near door fired with value OFF} \\
M025 & in walk-in closet & Motion & ON & \textit{After one hundred and twenty seconds, Motion sensor in walk-in closet fired with value ON} \\
M026 & in workspace / TV room & Motion & ON & \textit{After two seconds,  Motion sensor in workspace/TV room fired with value ON} \\
M020, M028 & in bedroom & Motion & ON & \textit{Motion sensor in bedroom fired with value ON at two hours three minute PM} \\
D002 & on coat cabinet near home entrance door & Door & OPEN & \textit{Door sensor on coat cabinet near home entrance door with value CLOSE at ten hours thirty minute PM} \\
T001 & in kitchen near stove & Temperature & 24 & \textit{After two seconds,  Temperature sensor in kitchen near stove with value twnety-four} \\
\bottomrule
\end{tabularx}
\end{adjustbox}
\caption{ Conversion table to build textual description using TDOST-Temporal variant for for a subset of sensors in CASAS-Milan dataset. } 
\label{tab:milan_tdost_temporal}

\end{table}

\subsubsection{Conversion process for  TDOST-LLM:}\label{appendix_tdost_llm}
In the TDOST-LLM version, a diverse set of text sentences encapsulating each sensor reading is generated using the pre-trained large language models. 
To show examples of TDOST generated by LLM, we randomly selected a sequence of "bed-to-toilet" activity triggers from CASAS Milan datasets and created TDOSTs using the floor plan in Fig. \ref{fig:milan_floorplan}.
In Tab. \ref{tab:milan_tdost_llm}, we present this sequence of randomly selected sensor triggers, the sensor reading representation used for prompting LLM, and the corresponding text sentences generated by LLM.

\begin{table}
\centering

\begin{tabularx}{\linewidth}{p{3cm}p{4cm}X}
\toprule

\textbf{Sensor Reading} & \textbf{Key Used in LLM Prompt} & \textbf{LLM Generated Sentences} \\ 
\midrule
M028 ON 21:29:11 & (Friday, Late Night, Motion, in bedroom, ON) & ["In the stillness of the late Friday night, the bedroom's motion sensor sprang to life, suggesting someone had just entered.", "The bedroom's still air was pierced by the activation of the motion sensor, recording a disturbance late on Friday night.", "An unseen guest seemed to have ventured into the bedroom late Friday night, as indicated by the motion sensor's alert."] \\
\midrule
M020 OFF 21:29:15 & (Friday, Late Night, Motion, in bedroom, OFF) & ["The bedroom was still and quiet, with no movement detected late on Friday night.", "As the clock struck deep into late night on Friday, the motion sensor in the bedroom remained inactive, signaling an undisturbed slumber.", "The absence of motion in the bedroom late at night on Friday indicated everyone had settled for the night."] \\
\midrule
M021 OFF 21:29:16 & (Friday, Late Night, Motion, in bedroom on bed, OFF) & ["After a brief stir, the bedroom bed motion sensor ceased its alert late on Friday night, suggesting the occupant had returned to rest.", "The motion on the bed came to a standstill late Friday night, as the individual may have found peace in the embrace of sleep.", "Following a momentary flurry of activity, the bedroom bed sensor quieted down, hinting at the sleeper's return to tranquility."] \\
\midrule
M025 ON 21:29:16 & (Friday, Late Night, Motion, in walk-in closet, ON) & ["A hushed motion was detected within the walk-in closet late at night on Friday, suggesting a discreet presence.", "The solitude of the walk-in closet was broken late Friday night as the motion sensor reported unexpected activity.", "Late on Friday night, the motion sensor inside the walk-in closet was triggered, hinting at some nocturnal shuffling."] \\
\midrule
M028 OFF 21:29:16 & (Friday, Late Night, Motion, in bedroom, OFF) & ["The bedroom was still and quiet, with no movement detected late on Friday night.", "As the clock struck deep into late night on Friday, the motion sensor in the bedroom remained inactive, signaling an undisturbed slumber.", "The absence of motion in the bedroom late at night on Friday indicated everyone had settled for the night."] \\
\midrule
M013 ON 21:29:20 & (Friday, Late Night, Motion, in bathroom near sink, ON) & ["In the solitude of late-night Friday, the bathroom sensor near the sink was triggered, suggesting a nocturnal visitor.", "The motion detector by the bathroom sink sprang to life, signaling an unexpected presence during the late hours of Friday.", "Activity was registered by the motion sensor in the bathroom next to the sink, piercing the late-night silence."] \\
\bottomrule

\end{tabularx}

\caption{ In this table, we show diverse sentences generated by LLM for a random sequence of sensor triggers depicting `Bed-to-toilet' activity in CASAS-Milan smart home.} 
\label{tab:milan_tdost_llm}
\end{table}

\subsubsection{Conversion map for TDOST-LLM+Temporal:}\label{appendix_tdost_llm_temp}

\begin{figure}[]
  \centering
  \includegraphics[width=0.9\linewidth]{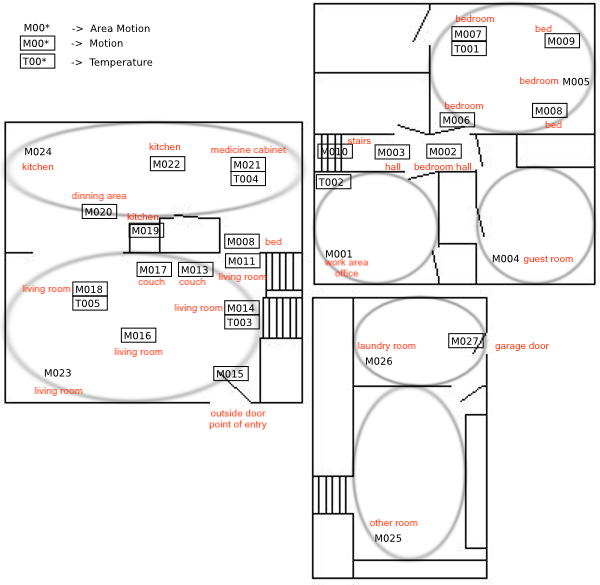} 
  \caption{Cairo sensor layout and floorplan (taken with permission from \cite{cook2012casas}) }
\label{fig:cairo_floorplan}
\end{figure}

The process followed for constructing text sentences in TDOST-LLM+Temporal is similar to TDOST-LLM except that we add  time information to each of the sentences generated by LLM.
The process is illustrated in Tab. \ref{tab:cairo_tdost_llm} where we randomly pick a sequence of sensor readings for "eating" activities from the Cairo dataset and show the final TDOST-LLM+Temporal sentences. 
For reference, the sensor arrangement in Cairo floorplan is depicted in Fig. \ref{fig:cairo_floorplan}

\begin{table}
\centering

\begin{tabularx}{\linewidth}{p{3cm}p{4cm}X}

\textbf{Sensor Reading} & \textbf{Key Used in LLM Prompt} & \textbf{LLM Generated Sentences +Temporal} \\ 

\toprule
M012 ON 06:27:36 & (Wednesday, Early Morning, Motion, in kitchen, ON) & "The kitchen motion sensor registered activity early Wednesday morning, perhaps a sign of an early bird preparing breakfast at six hours and twenty-seven seconds AM" \\
\hline
M016 OFF 06:27:36 & (Wednesday, Early Morning, Motion, in living room, OFF) & "After zero seconds, the early morning bustle in the living room has subsided, with the motion sensor now indicating no further movement on Wednesday." \\
\midrule
M011 OFF 06:27:39 & (Wednesday, Early Morning, Motion, in living room near bottom of stairs, OFF) & "After two seconds, the early morning bustle slowed down as the motion sensor in the living room near the stairs ceased detecting movement on Wednesday." \\
\midrule
M020 ON 06:27:39 & (Wednesday, Early Morning, Motion, in dinning area near kitchen, ON) & "After zero seconds, an early riser stirred in the dining area close to the kitchen, triggering the motion sensor on this Wednesday morning." \\
\midrule
M019 ON 06:27:39 & (Wednesday, Early Morning, Motion, in kitchen, ON) & "After zero seconds, the kitchen motion sensor registered activity early Wednesday morning, perhaps a sign of an early bird preparing breakfast." \\
\midrule
M018 OFF 06:27:40 & (Wednesday, Early Morning, Motion, in living room, OFF) & "After one  seconds, the early morning bustle in the living room has subsided, with the motion sensor now indicating no further movement on Wednesday."  \\
\midrule
M012 OFF 06:27:40 & (Wednesday, Early Morning, Motion, in kitchen, OFF) & "After zero seconds,  culinary preparations came to a pause in the kitchen on Wednesday morning, reflected by the motion sensor's transition to an inactive state" \\
\midrule
M013 ON 06:27:41 & (Wednesday, Early Morning, Motion, near couch in living room, ON) & "After one seconds, the presence of someone or something near the couch in the living room was detected by the motion sensor, as it switched on early Wednesday morning." \\
\bottomrule
\end{tabularx}

\caption{ In this table, we show LLM-generated text sentences along with  temporal information used in  TDOST-LLM+Temporal version for a random sequence of sensor triggers depicting `Eating' activity in CASAS-Cairo smart home.} 
\label{tab:cairo_tdost_llm}

\end{table}

\end{document}